\documentclass[a4paper,fleqn]{cas-dc}

\usepackage[authoryear]{natbib}
\usepackage{subfigure}
\makeatletter
\renewcommand{\@thesubfigure}{\hskip\subfiglabelskip}
\makeatother

\def\tsc#1{\csdef{#1}{\textsc{\lowercase{#1}}\xspace}}
\tsc{WGM}
\tsc{QE}

\begin{document}
\let\WriteBookmarks\relax
\def\floatpagepagefraction{1}
\def\textpagefraction{.001}

\shorttitle{Texture Sensitive Semantic Segmentation and Advanced Style Transfer Learning Strategy}    

\shortauthors{Z. Yang et al.}

\title[mode = title]{An easy zero-shot learning combination: Texture Sensitive Semantic Segmentation IceHrNet and Advanced Style Transfer Learning Strategy}

\ExplSyntaxOn
\keys_set:nn { stm / mktitle } { nologo }
\ExplSyntaxOff

\author[1]{Zhiyong Yang}[orcid=0009-0008-1511-9203]
\ead{yzy_hohai@hhu.edu.cn}
\ead[url]{https://github.com/PL23K}
\credit{Conceptualization of this study, Methodology, Software, Writing - Original draft preparation}

\author[1]{Yuelong Zhu}
\ead{ylzhu@hhu.edu.cn}
\credit{Conceptualization of this study, Methodology}

\author[1]{Xiaoqin Zeng}
\ead{xzeng@hhu.edu.cn}
\credit{Conceptualization of this study, Supervision}

\author[2]{Jun Zong}
\ead{121484530@qq.com}
\credit{Resources, Funding acquisition}

\author[1,3]{Xiuheng Liu}
\ead{liuxiuheng@hhu.edu.cn}
\credit{Validation}

\author[2]{Ran Tao}
\ead{2630411035@qq.com}
\credit{Visualization, Data curation}

\author[1,3]{Xiaofei Cong}
\ead{congxiaofei@hhu.edu.cn}
\credit{Investigation}

\author[1]{Yufeng Yu}
\ead{yfyu@hhu.edu.cn}
\cormark[1] 
\credit{Conceptualization of this study, Project administration, Writing- Reviewing and Editing}

\address[1]{College of Computer and Information Engineering, Hohai University, Nanjing 211100, China}
\address[2]{Nanjing Research Institute of Hydrology and Water Conservation Automation, Ministry of Water Resources 95\#Tiexinqiao Rd., Nanjing 210012, China}
\address[3]{Nanjing Zhongyu Intelligent Water Conservancy Research Institute Co., Ltd, Nanjing 210012, China}

\cortext[1]{Corresponding author:yfyu@hhu.edu.cn;Tel.:+86-13951670067} 

\begin{abstract}
We proposed an easy method of Zero-Shot semantic segmentation by using style transfer. In this case, we successfully used a medical imaging dataset (Blood Cell Imagery) to train a model for river ice semantic segmentation. First, we built a river ice semantic segmentation dataset IPC\_RI\_SEG using a fixed camera and covering the entire ice melting process of the river. Second, a high-resolution texture fusion semantic segmentation network named IceHrNet is proposed. The network used HRNet as the backbone and added ASPP and Decoder segmentation heads to retain low-level texture features for fine semantic segmentation. Finally, a simple and effective advanced style transfer learning strategy was proposed, which can perform zero-shot transfer learning based on cross-domain semantic segmentation datasets, achieving a practical effect of 87\% mIoU for semantic segmentation of river ice without target training dataset (25\% mIoU for None Stylized, 65\% mIoU for Conventional Stylized, our strategy improved by 22\%). Experiments showed that the IceHrNet outperformed the state-of-the-art methods on the texture-focused dataset IPC\_RI\_SEG, and achieved an excellent result on the shape-focused river ice datasets. In zero-shot transfer learning, IceHrNet achieved an increase of 2 percentage points compared to other methods. Our code and model are published on \href{https://github.com/PL23K/IceHrNet}{https://github.com/PL23K/IceHrNet}.
\end{abstract}

\begin{graphicalabstract}
	\includegraphics[width=\linewidth]{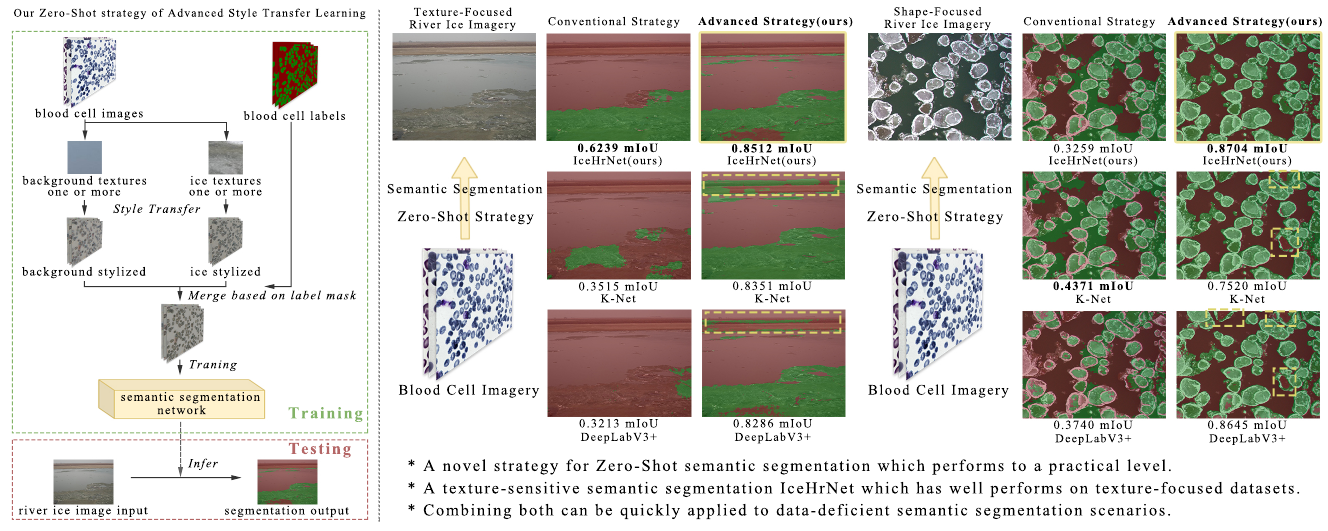}
\end{graphicalabstract}

\begin{highlights}
\item A novel strategy of Zero-Shot semantic segmentation performs to a practical level.
\item Texture-sensitive semantic segmentation network: IceHrNet.
\item This combination could be quickly applied to data-deficient semantic segmentation.
\item IPC\_RI\_SEG dataset covers the entire ice melting process for ice conditions research.
\end{highlights}


\maketitle

\section{Introduction}
River ice floods often cause huge losses downstream \cite{Yang_Pavelsky_Allen_2020}, it is important to establish a hydrological ice flood warning. River ice semantic segmentation and stage estimation methods can achieve ice flood early warning. Therefore, it is necessary to monitor the formation and melting status of river ice in real time through computer vision technology. Based on river ice semantic segmentation technology, the pixel distribution of ice and water in river camera images is identified, further extracting high semantic information such as ice cover density, ice drift speed, ice cover distribution, and changing process.

In general, many researchers used high spatial resolution remote sensing images (Satellite Imagery) \cite{Zhang_Yue_Han_Li_Yuan_Fan_Zhang_2021, Li_Li_Wang_Hao_2023}, unmanned aerial vehicle aerial photography (UAV Imagery) \cite{Singh_Kalke_Loewen_Ray_2020, Zhang_Zhao_Ran_Xing_Wang_Lan_Yin_He_Liu_Zhang_et_al_2023}, and images captured by fixed position cameras (Fixed Camera Imagery) \cite{Pei_She_Loewen_2022} to analyze river ice conditions through computer vision models. Satellite Imagery advantages wide range and UAV Imagery advantages not limited by time and location, while Fixed Camera Imagery advantages fixed viewing angle and real-time monitoring. It is more suitable for detailed analysis and changing stage identification of river ice. In this paper, we are using Fixed Camera Imagery to research river ice condition recognition.

\textbf{River Ice Semantic Segmentation}
\begin{figure}[h]
	\centering
	\subfigure[(a)]{
		\begin{minipage}[b]{0.44\linewidth}
			\includegraphics[width=0.48\linewidth]{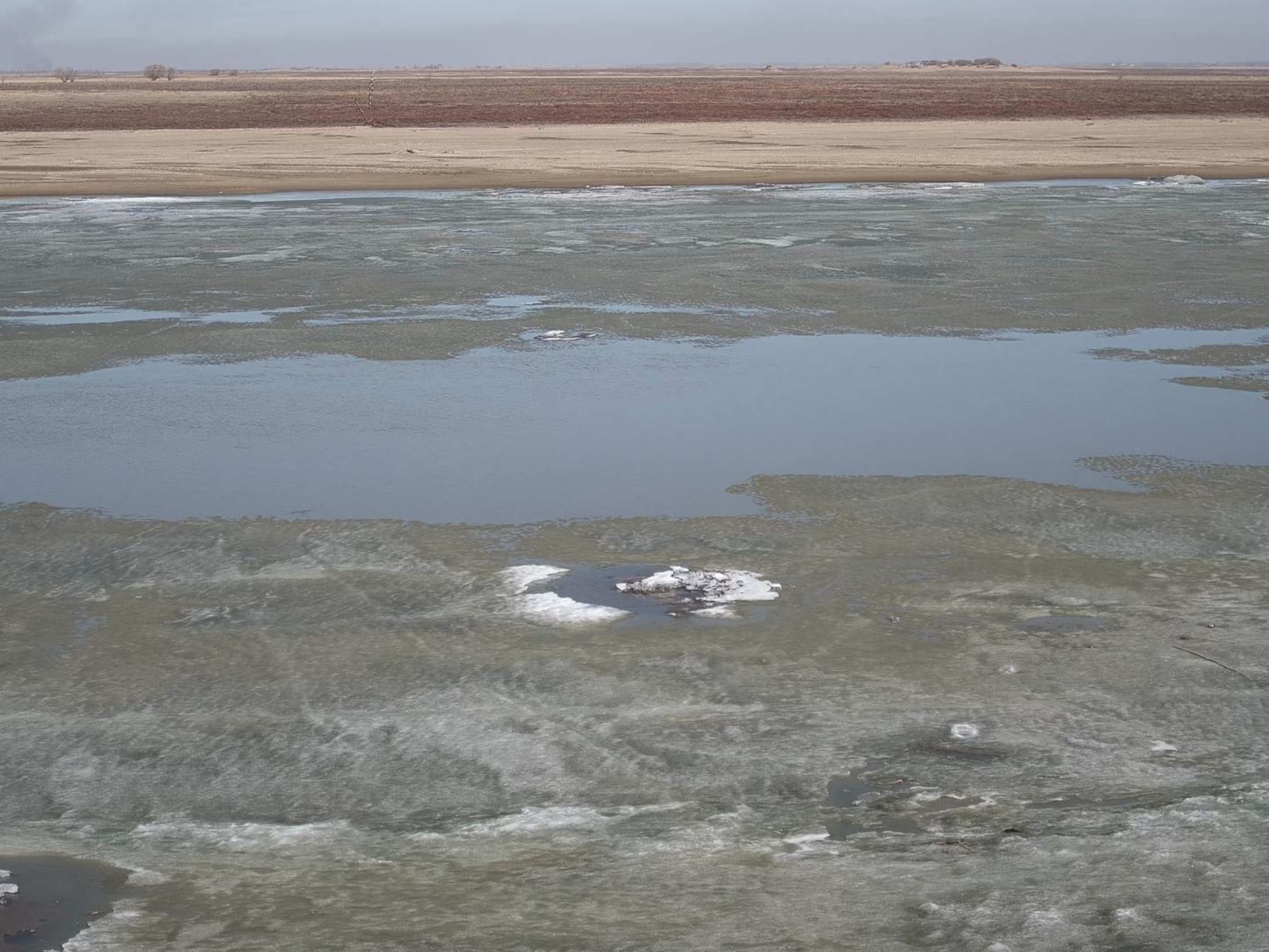}
			\includegraphics[width=0.48\linewidth]{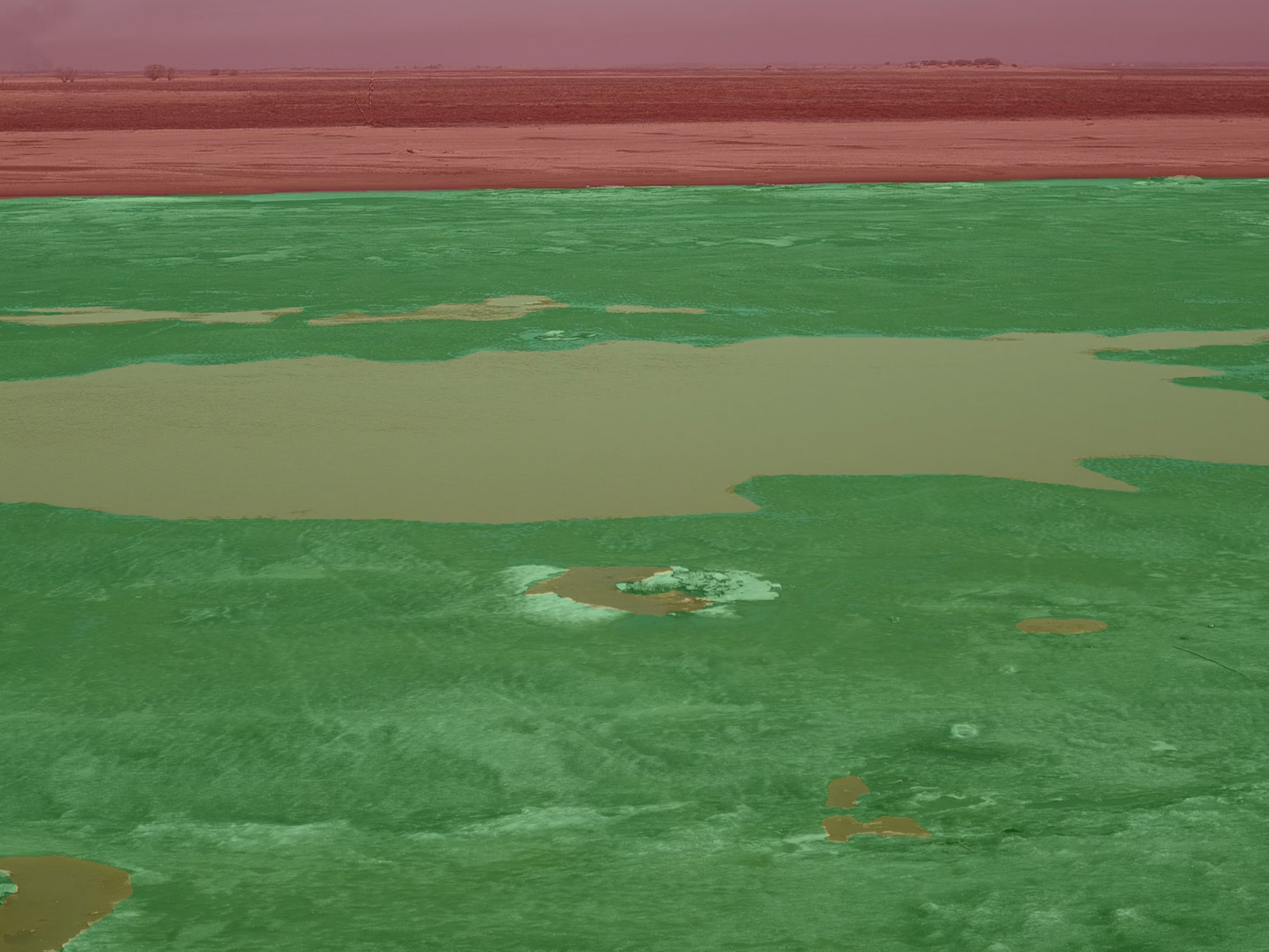}
		\end{minipage}
	}
	\subfigure[(b)]{
		\begin{minipage}[b]{0.44\linewidth}
			\includegraphics[width=0.48\linewidth]{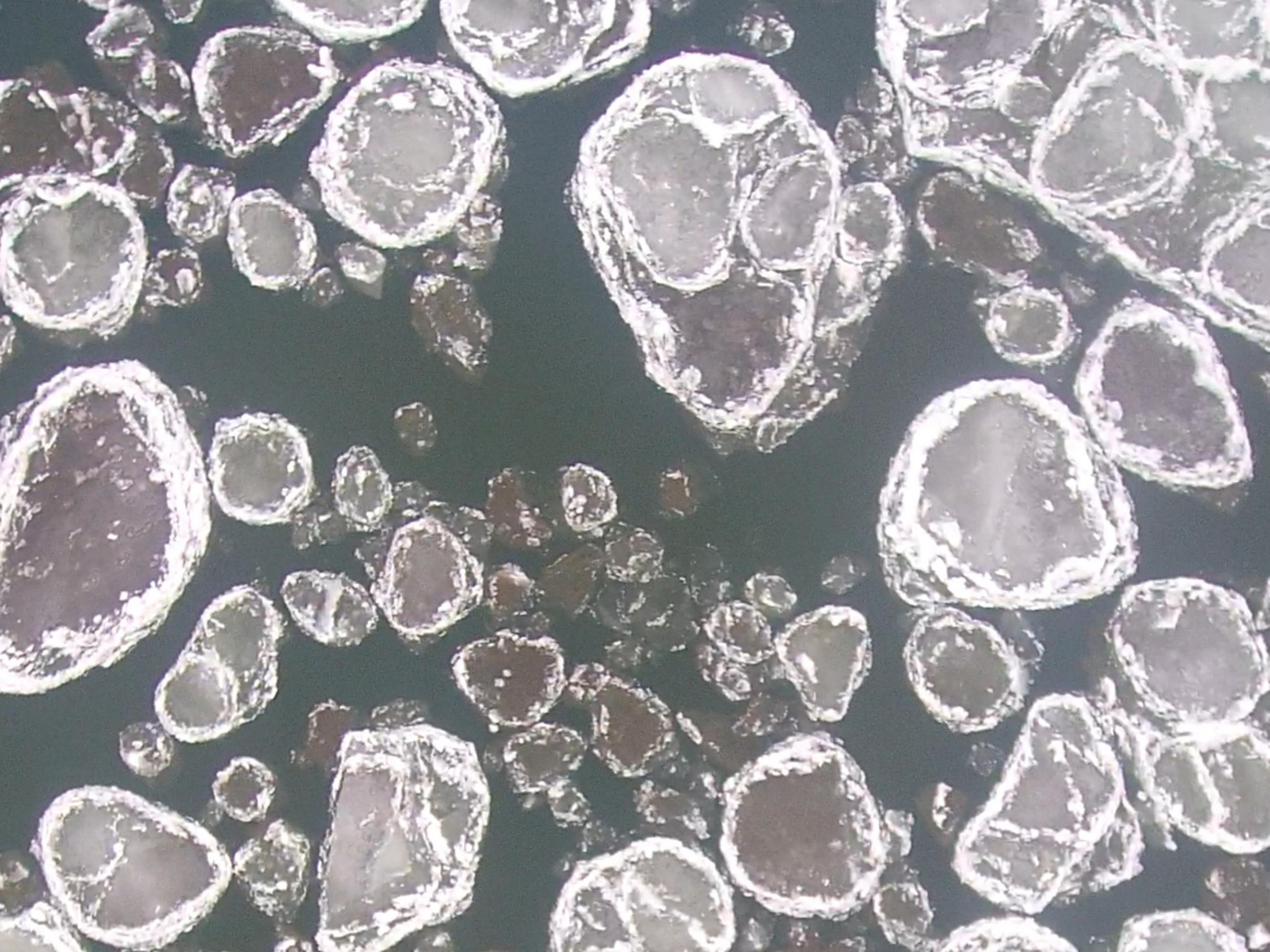}
			\includegraphics[width=0.48\linewidth]{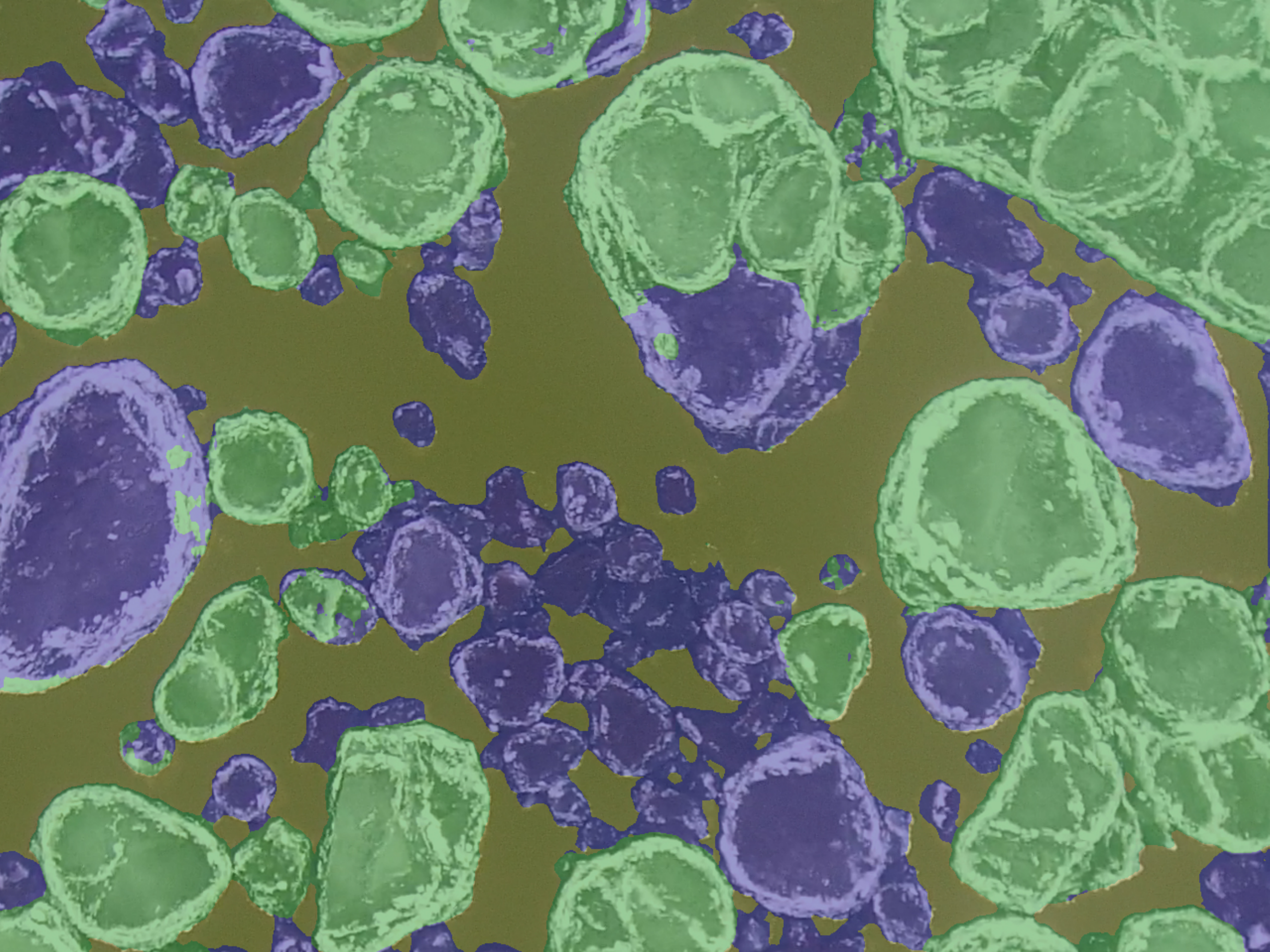}
		\end{minipage}
	}
	\caption{The examples of semantic segmentation. Identify for each pixel in the image which category it belongs to, water, background, ice, or anchor ice. (a) River ice from Fixed Camera Imagery. (b) River ice From UAV Imagery.}\label{Figure1}
\end{figure}

Semantic segmentation is a very effective technology that contribute to understand river ice conditions (see \textbf{Figure \ref{Figure1}}). Early researchers used traditional machine learning methods to analyze river ice images and obtain approximate river ice distribution \cite{Unterschultz_van_der_Sanden_Hicks_2009, Ansari_Rennie_Seidou_Malenchak_Zare_2017, Kalke_Loewen_2018, Daigle_Berube_Bergeron_Matte_2013}. With the development of deep convolutional neural networks, the automatic monitoring of river ice conditions is also continuously advancing. \cite{Singh_Kalke_Loewen_Ray_2020} published a river ice semantic segmentation dataset based on UAV imagery online and conducted comparative experiments using various deep convolutional neural networks, contributing to subsequent research in this field. Immediately afterward, Zhang et al. did  three consecutive works on semantic segmentation of river ice based on drone images \cite{Zhang_Jin_Lan_Li_Fan_Wang_Yu_Zhang_2020, Zhang_Zhou_Jin_Wang_Fan_Wang_Zhang_2021, Zhang_Zhao_Ran_Xing_Wang_Lan_Yin_He_Liu_Zhang_et_al_2023} . \cite{Ansari_Rennie_Clark_Seidou_2021} proposed the IceMaskNet for instance segmentation of river ice. Similar to the work of this paper, \cite{Pei_She_Loewen_2022} collected river ice images and performed semantic segmentation of river surface ice. They conducted long-term monitoring of the river through fixed cameras. Their main research content was surface ice concentration and ice pan characteristics in different long periods.

\textbf{Style transfer learning}

Style Transfer and Style Transfer Learning. \textbf{Style Transfer} includes text style transfer and neural style transfer, and this paper refers to neural style transfer. \textbf{Style Transfer Learning} is explained in this paper as transfer learning based on style transferring, aiming at cross-domain transfer learning research.

\begin{figure}[h]
	\centering
	\subfigure[(a)]{
		\includegraphics[width=0.44\linewidth]{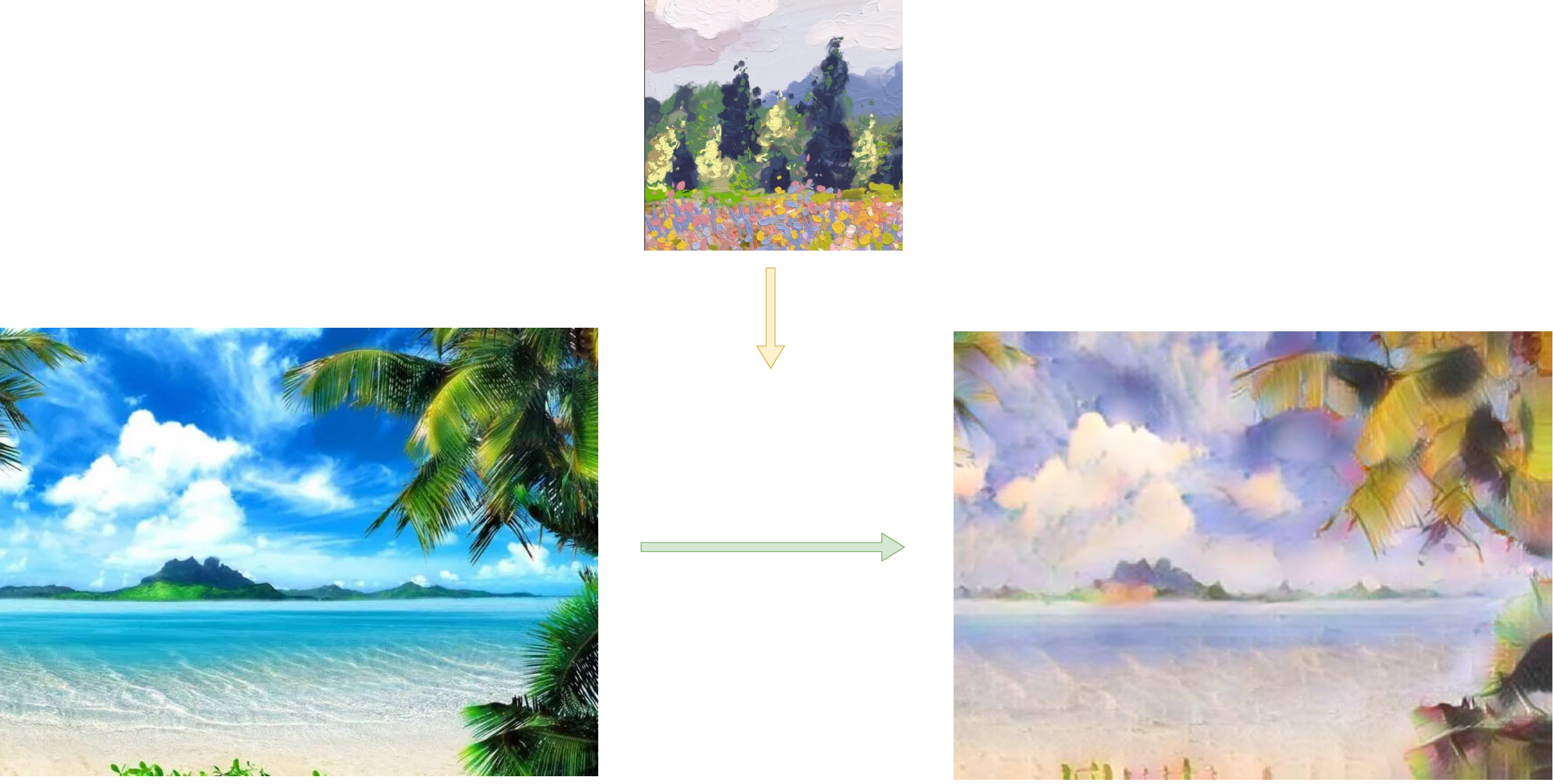}
	}
	\subfigure[(b)]{
		\includegraphics[width=0.44\linewidth]{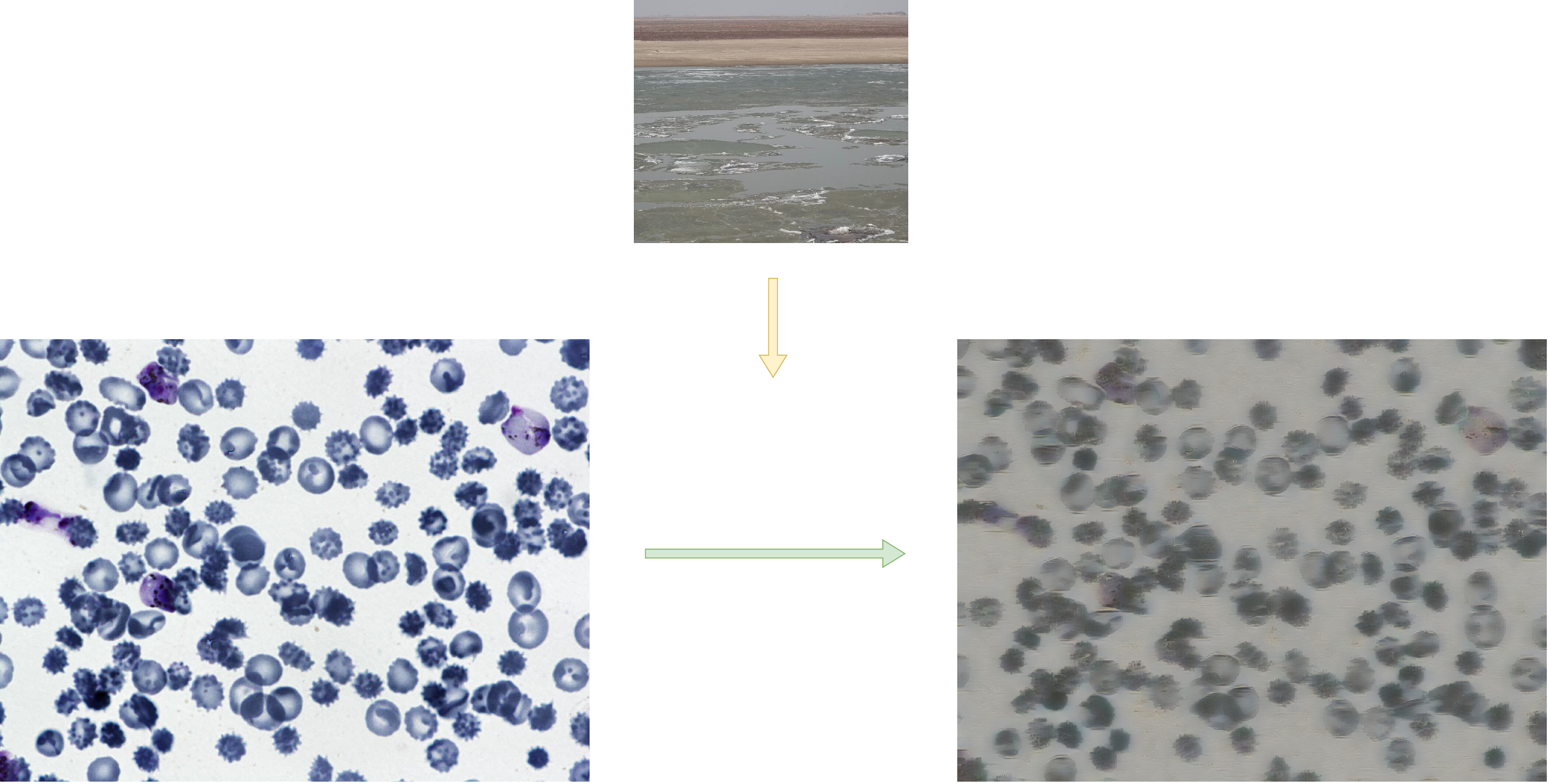}
	}
	\caption{The examples of style transfer. (a) A landscape image is stylized into an oil painting. (b) A medical cell image is stylized into a river ice image.}\label{Figure2}
\end{figure}

Style transfer is the process of using CNN to fuse the semantic content of an image with different styles. The research of \cite{Gatys_Ecker_Bethge_2016} successfully used convolutional neural networks for style transfer for the first time. Since then, neural style transfer has quickly become a very popular topic, attracting the attention of many scholars, and many interesting works have appeared. Such as face replacement \cite{Karras_Laine_Aila_2019}, oil painting transfer \cite{Liu_Lin_He_Li_Deng_Li_Ding_Wang_2021}, and so on (see \textbf{Figure \ref{Figure2}}).

Style transfer learning was first used as a data augmentation method in other deep learning tasks by \cite{Perez_Wang_2017}. \cite{Jackson_Atapour_Abarghouei_Bonner_Breckon_Obara_2019} studied the impact of style augmentation on domain transfer tasks. Later, the combination of semantic segmentation and style transfer evolved into two branches. One is to use the characteristics of semantic segmentation to improve the effect of style transfer \cite{Benitez_Garcia_Shimoda_Matsuo_Yanai_2020, Lin_Wang_Chen_Ma_Xie_Xing_Zhao_Song_2021, Huang_Yang_Tang_Zhuang_Hou_Tan_Dananjayan_He_Guo_Luo_2021}, and the second type is to use the characteristics of style transfer to improve the effect of semantic segmentation\cite{Song_Xu_Zhang_2020,Kang_Zang_Cao_2021, Marsden_Wiewel_Dobler_Yang_Yang_2022,Shan_Yin_Gao_Liang_Ma_Guo_2022, Nam_Nguyen_Dieu_Visani_Nguyen_Sang_2022}.

Related to this paper, \cite{Li_Ye_Cao_Hou_Yang_2021} used style transfer to achieve zero-shot transfer in target recognition in ocean-side-scan sonar images. \cite{Kline_2021} proposed a method of changing image intensity and texture to improve performance on the task of segmenting the kidneys in patients affected by polycystic kidney disease. \cite{Zhao_Wei_Lu_Bai_Zhao_Chen_Hu_2023} proposed a cross-domain target detection model based on style transfer contrastive learning. This model is used to apply source domain data to generate new samples of the target domain style, thereby achieving cross-domain target detection. The above style transfer learning strategies are based on global-image style transfer and do not effectively utilize source domain category annotation information.

\textbf{Our Works}

\begin{figure*}[htbp]
	\centering
	\subfigure[(a) ice freezing period]{\includegraphics[width=0.18\linewidth]{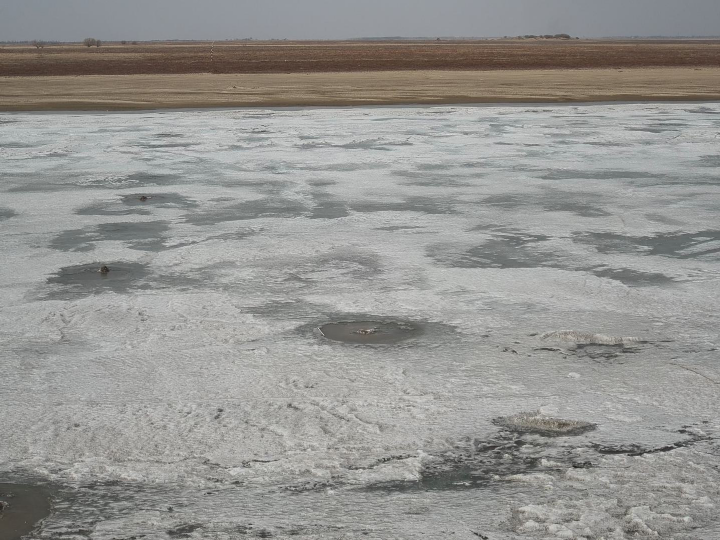}}
	\subfigure[(b) ice melting period]{\includegraphics[width=0.18\linewidth]{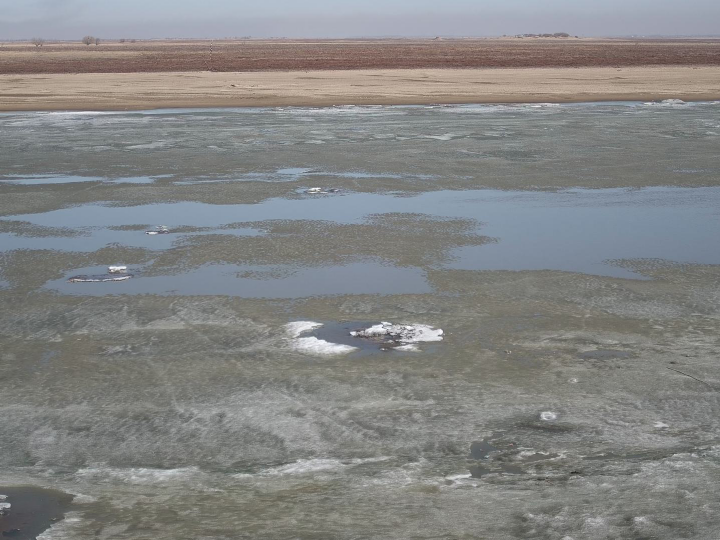}}
	\subfigure[(c) ice drifting period]{\includegraphics[width=0.18\linewidth]{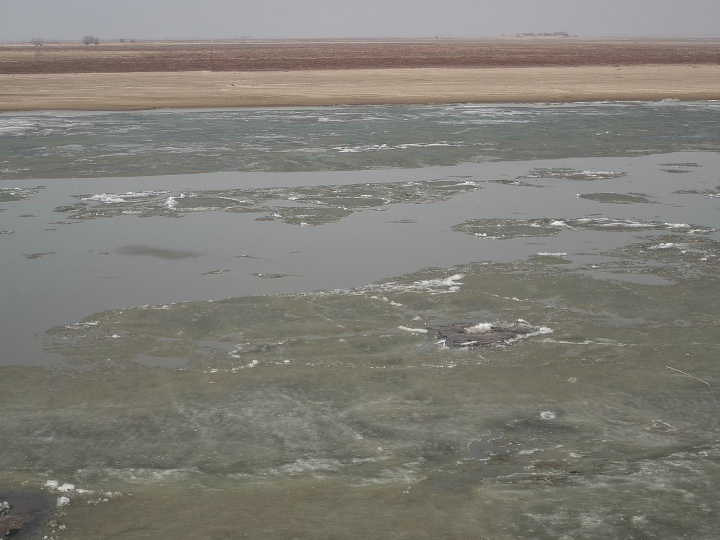}}
	\subfigure[(d) melt ending period]{\includegraphics[width=0.18\linewidth]{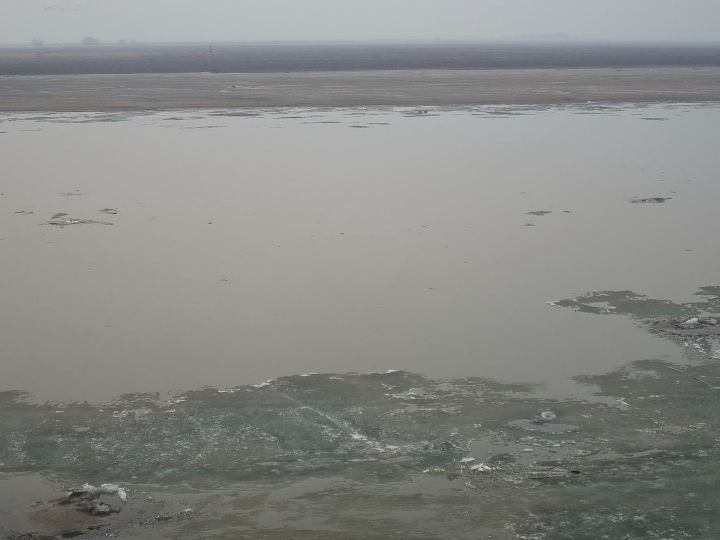}}
	\subfigure[(e) night]{\includegraphics[width=0.18\linewidth]{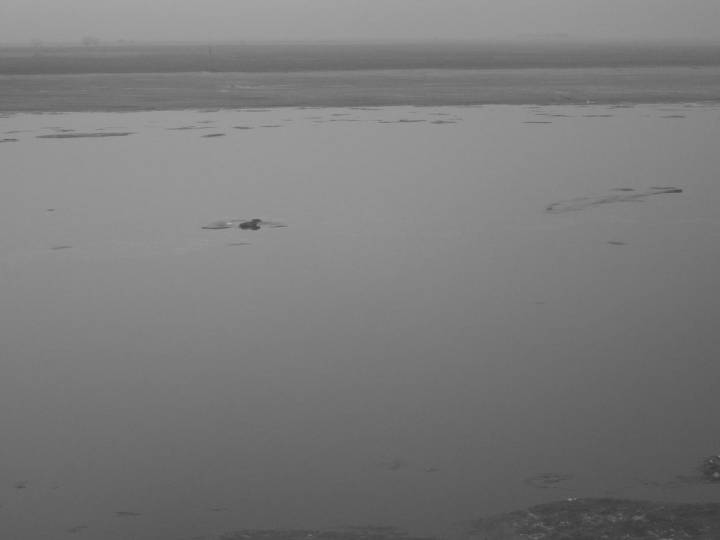}}
	\caption{Five stages of our river ice dataset. (a)ice freezing period; (b)ice melting period (c)ice drifting period (d)melt ending period (e)night.}\label{Figure3}
\end{figure*}

This paper explored the deep convolutional neural network semantic segmentation of river surface ice on fixed camera imagery, which extracts pixels of river ice from the camera image to provide basic information support for downstream river ice analysis tasks. In the real-time monitoring task of ice drift speed, semantic segmentation on fixed camera imagery is necessary. However, the main challenges currently faced in the river ice semantic segmentation task are:
\begin{enumerate}
	\item The river ice in different regions and river sections is different;
	\item Lacking dataset, high annotation cost, and diversity;
	\item Difficulty in segmenting fragmented ice in ice-water mixed during the melting stage.
\end{enumerate}

To address the above issues, we built a new river ice semantic segmentation dataset, called IPC\_RI\_SEG, which is collected from fixed position network cameras, contains 113 images with detailed annotation, and covers the entire ice melting process (see \textbf{Figure \ref{Figure3}}). Our dataset supplemented the shortcomings of river ice vision research in fixed camera imagery datasets. We proposed a high-resolution texture fusion deep convolutional neural network for semantic segmentation named IceHrNet, which used HRNet \cite{Sun_Xiao_Liu_Wang_2019} as the backbone and adds ASPP \cite{Chen_Zhu_Papandreou_Schroff_Adam_2018} and Decoder segmentation heads to retain low-level texture features for fine semantic segmentation. In zero-shot transfer learning based on style transferring, it has good memory ability for texture features.

Although the IPC\_RI\_SEG dataset was built, the data was still limited and cannot be well generalized to other scenarios. We were inspired by image style transfer techniques \cite{Gatys_Ecker_Bethge_2016}. A novel strategy was proposed to achieve cross-domain zero-shot semantic segmentation, training the source domain dataset that style transferred by label category using the target domain dataset. This strategy performsed to a practical level. This could reduce the overhead of semantic segmentation dataset collection in specific scenarios and quickly complete the segmentation task.

The main contributions of this paper are as follows:

\begin{enumerate}
\item We proposed a novel simple and effective strategy of \textbf{Advanced Style Transfer Learning}, which could achieve cross-domain zero-shot semantic segmentation, training the source domain dataset which was style transferred by the target domain dataset label categories. In the case of this paper, we use medical imaging datasets as source domain datasets for transfer learning and apply them to the semantic segmentation of river ice datasets, the result performs to a practical level. Our strategy of advanced style transfer learning is very suitable for the semantic segmentation of irregularly shaped objects.
\item We proposed a high-resolution texture fusion deep convolutional neural network for semantic segmentation named \textbf{IceHrNet}, which used HRNet as the backbone and adds ASPP and Decoder segmentation heads to retain low-level texture features for fine semantic segmentation. In zero-shot transfer learning based on style transferring, it has good memory ability for texture features.
\item We built a river ice semantic segmentation dataset named \textbf{IPC\_RI\_SEG} based on fixed camera imagery, which supplemented the shortcomings of river ice visual research in fixed camera imagery datasets. This dataset also contained the entire process of river ice in the freezing, melting, and receding stages, which is helpful for research on early warning of river ice conditions.
\end{enumerate}  

\section{Methods}
\subsection{Texture Sensitive Network IceHrNet}

\begin{figure}[h]
	\centering
	\includegraphics[width=0.98\linewidth]{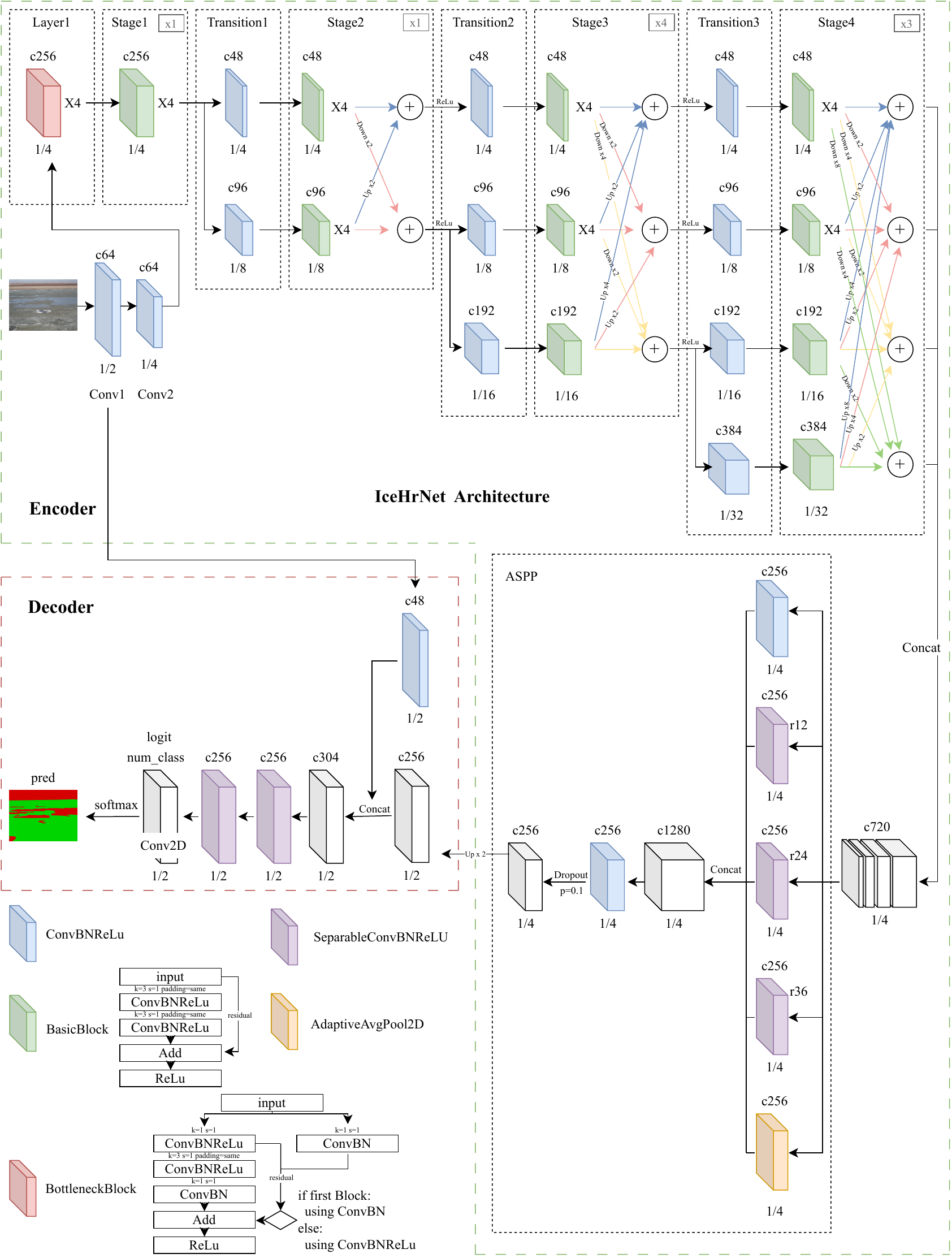}
	\caption{The network architecture of IceHrNet. As you can see, “c64” and “c256” represent 64 channels and 256 channels, etc. “1/4” represents a quarter of the original image size. “X4” represents repeat 4 times.}\label{Figure4}
\end{figure}

To address the problem of accurately segmenting fragmented ice and boundaries in ice-water mixing in this research work, it is necessary to maintain low-level texture features in the deep convolutional neural network to prevent the loss of texture details. Therefore, we proposed the IceHrNet deep convolutional neural network, which is a semantic segmentation network based on the HRNet \cite{Sun_Xiao_Liu_Wang_2019} high-resolution network as the backbone and adds ASPP \cite{Chen_Zhu_Papandreou_Schroff_Adam_2018} and Decoder segmentation head to retain low-level texture features for fine semantic segmentation. The network architecture of IceHrNet is shown in \textbf{Figure \ref{Figure4}}.

IceHrNet is composed of an Encoder and a Decoder. The Encoder is a feature extractor that includes an HRNet high-resolution feature extraction module and an ASPP module. The Decoder is a feature decoder that fuses the shallow features with the high-level features of the Encoder to obtain prediction results. The network Encoder fully extracts multi-scale features while maintaining high-resolution features. The Decoder once again incorporates shallow features, so the details of the input image could be better distinguished. There are two options for fusing shallow features, the first is to fuse the feature map of Conv1, and the second is to fuse the feature map of Conv2. We used the shallower Conv1.

Specifically, the input image first passes through two convolutional layers, namely Conv1 and Conv2, for two downsampling shallow feature extractions. It should be noted that these two shallow features are used in the Decoder. Then, the feature map enters the HRNet backbone with 1 layer, 4 stages, and 3 transitions. Then output a feature map whose size is 1/4 of the original image and the number of channels is 720. The feature map enters an ASPP module, to reduce the channels to 256. This 256-channel feature map is the final output of the Encoder. The Decoder is relatively simple. The Decoder fuses the shallow features in the Encoder with the final output features and then obtains the prediction map. There are two options for the fusion of shallow features, the first is to fuse the feature map of Conv1, and the second is to fuse the feature map of Conv2. We found through experiments that in the river ice dataset, the result of the fusion of Conv1 will be better than the result of Conv2, as shown in \textbf{Table \ref{tbl5}}, Therefore, the final IceHrNet is determined to choose to fuse the Conv1 feature map.

\subsection{Advanced Style Transfer Learning}

\begin{figure*}[h]
	\centering
	\includegraphics[width=0.98\linewidth]{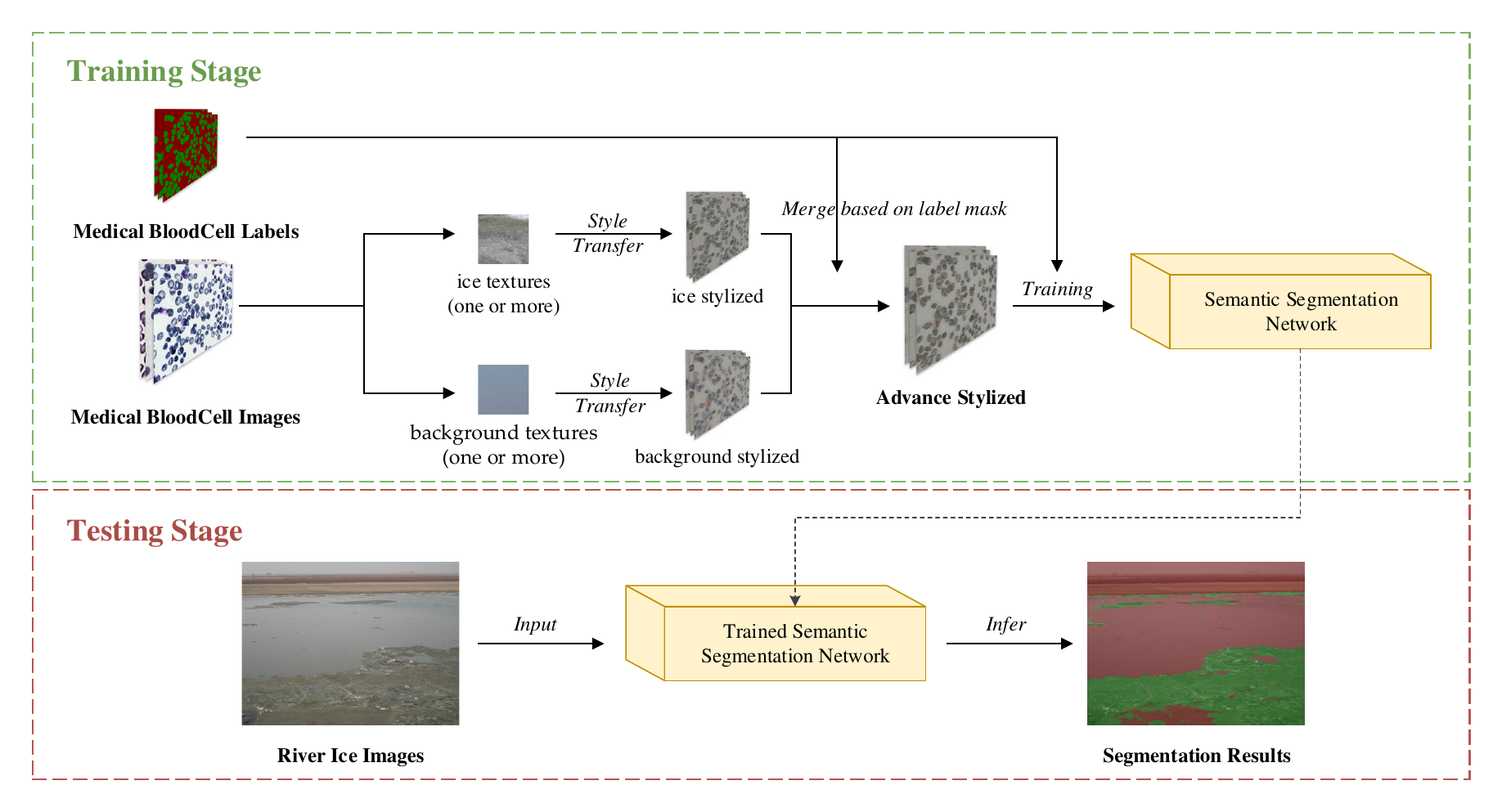}
	\caption{The strategy of Advanced Style Transfer Learning.}\label{Figure5}
\end{figure*}

The difference between our strategy and previous style transfer learning \cite{Li_Ye_Cao_Hou_Yang_2021, Kline_2021, Zhao_Wei_Lu_Bai_Zhao_Chen_Hu_2023} is that their strategies all use a neural style network to convert the global image of the source domain into the style of the target domain image. In this paper, we refer to these strategies as \textbf{Global Style Transfer Learning} or \textbf{Conventional Style Transfer Learning}. Our strategy is to use the labels of the source domain dataset to perform style conversion on the source domain images separately by category, and then merge the converted style maps according to the category position to generate target style samples with class discrimination capabilities. In this paper, we refer to our strategy as \textbf{Advanced Style Transfer Learning} (see \textbf{Figure \ref{Figure5}}). The case of this paper is to apply the medical imaging dataset transfer to river ice semantic segmentation.

The advanced style transfer learning strategy is very simple and easy to implement. First, perform a style transferring on the source domain dataset (medical imaging blood cell dataset) by the target domain dataset (river ice dataset) label categories. The style transfer method could be freely chosen. It is recommended to select from the literature \cite{Lin_Ma_Li_He_Li_Ding_Wang_Li_Gao_2021} according to your scenario. Secondly, according to the semantic labels of the source domain dataset, the images with corresponding category styles are extracted by pixels, and then all the extraction results are merged according to the pixel positions of the semantic labels, thereby obtaining improved stylized images with distinct categories. Finally, select a semantic segmentation network suitable for the target task and train using the stylized dataset. The resulting model can be directly applied to semantic segmentation in the target domain.

Note that the number of categories in the source domain dataset needs to be greater than or equal to the number of categories in the target domain. The target domain's style image (category textures) could be manually intercepted from one or more images to cover the texture of the target domain category.

\section{Results}

This experiment uses the deep learning framework PaddlePaddle2.5.1, PaddleSeg2.8, Python3.8.17, CUDA11.7, and CUDNN8.4.1.50. The experimental equipment is an NVIDIA GPU GeForce RTX 3090, 24GB VRAM, 32GB RAM, i7 10700 CPU, and CentOS7 operating system.

\subsection{Datasets Collection}

River ice datasets that are publicly available on the Internet and have semantic segmentation labels include the UAV Imagery dataset \cite{ebax-1h44-19} and Satellite Imagery dataset \cite{Canadian_Ice_Service_2009} . Currently, we have not found a downloadable public river ice dataset based on fixed camera imagery with semantic segmentation labels. We download and use the above two types of river ice datasets to supplement the research work on the river ice semantic segmentation task(see \textbf{Table \ref{tbl1}}).

\begin{table}[h]
	\centering
	\caption{River ice dataset used in this paper}\label{tbl1}
	\begin{tabular*}{\tblwidth}{@{}p{3cm} c p{3.2cm}@{}}
		\toprule
		Type & Images & Source \\
		\midrule
		Fixed Camera Imagery & 113 & \textbf{IPC\_RI\_SEG(Ours)} \\
		UAV Imagery & 50 & \multicolumn{1}{m{3.2cm}}{Alberta River Ice Segmentation Dataset} \\
		Satellite Imagery & 3392 & \multicolumn{1}{m{3.2cm}}{Canadian Ice Service Arctic Regional Sea Ice Charts in SIGRID-3 Format, Version 1} \\
		\bottomrule
	\end{tabular*}
\end{table}

\begin{figure}[h]
	\centering
	\subfigure[(a)Image]{\includegraphics[width=0.22\linewidth]{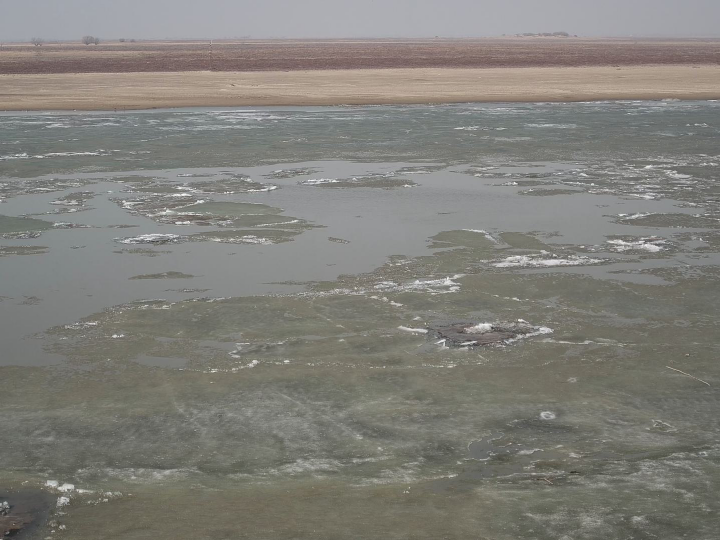}}
	\subfigure[(b)Label]{\includegraphics[width=0.22\linewidth]{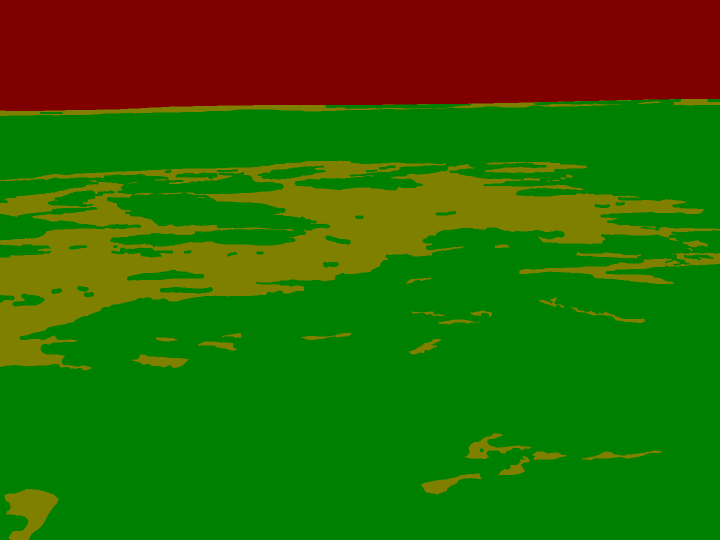}}
	\subfigure[(a)Image]{\includegraphics[width=0.22\linewidth]{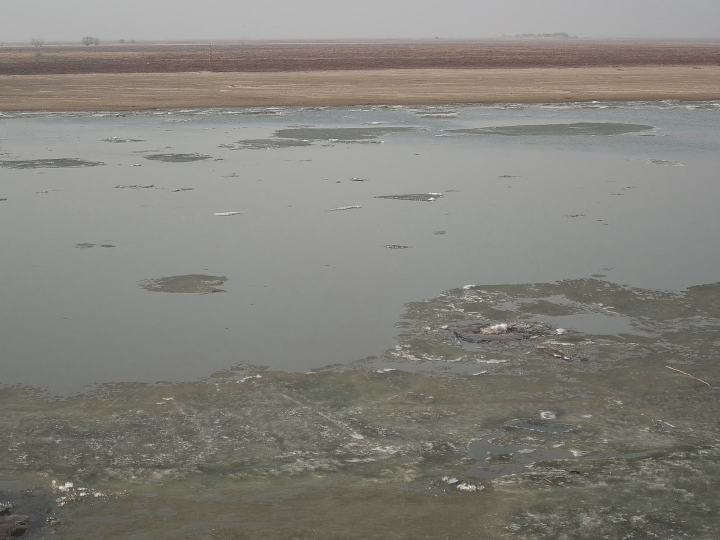}}
	\subfigure[(b)Label]{\includegraphics[width=0.22\linewidth]{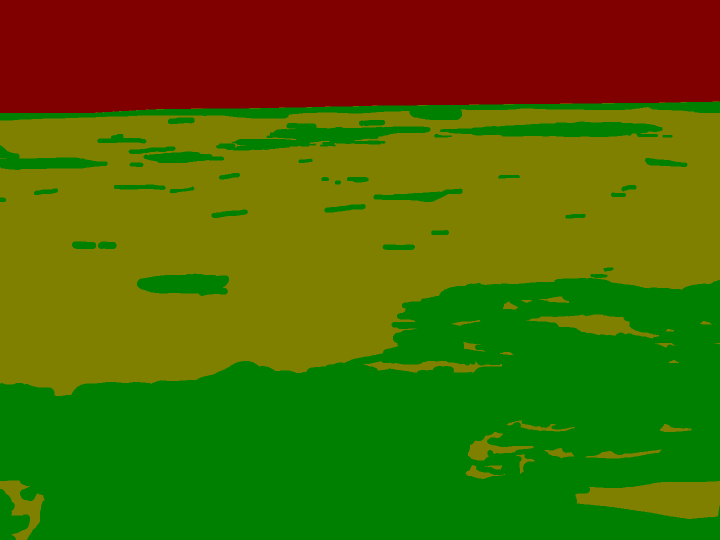}}
	\caption{Partial samples from our dataset. (a) the original images. (b) the labels for the original image.}\label{Figure6}
\end{figure}

\textbf{Fixed Camera Imagery (Ours)}. Our work mainly uses fixed camera monitoring to perform accurate semantic segmentation of river ice during the melting process, to provide data support for subsequent analysis of ice density and ice speed. Therefore, we chose a river in the plateau area of northern China. When the river melted, we used a fixed-position network camera on the river bank to record the entire ice-melting process. The ice melting started at 02:00 pm Beijing time and ended at 05:00 pm. A total of 43 video clips were saved (video memory size 43GB). We divide the ice melting process into five stages, namely ice freezing period, ice melting period, ice drifting period, melt ending period, or night. We extracted a total of 113 samples from five stages based on the characteristics of each stage and river ice conditions. We named the dataset IPC\_RI\_SEG. The images are shown in \textbf{Table \ref{tbl2}}, and Fine semantic annotation is provided, as shown in \textbf{Figure \ref{Figure6}}.

\begin{table}[h]
	\centering
	\caption{The number of images in each stage}\label{tbl2}
	\begin{tabular*}{\tblwidth}{@{}p{0.8cm} c c@{}}
		\toprule
		No. & The Stage of Ice Melting & The Number of Images \\
		\midrule
		1 & ice freezing period & 3 \\
		2 & ice melting period & 25 \\
		3 & ice drifting period & 48 \\
		4 & melt ending period & 24 \\
		5 & night & 25 \\
		\hline
		& \textbf{Total} & \textbf{113} \\
		\bottomrule
	\end{tabular*}
\end{table}

\begin{figure}[h]
	\centering
	\subfigure[(a) Original]{
		\begin{minipage}[b]{0.32\linewidth}
			\includegraphics[width=0.48\linewidth]{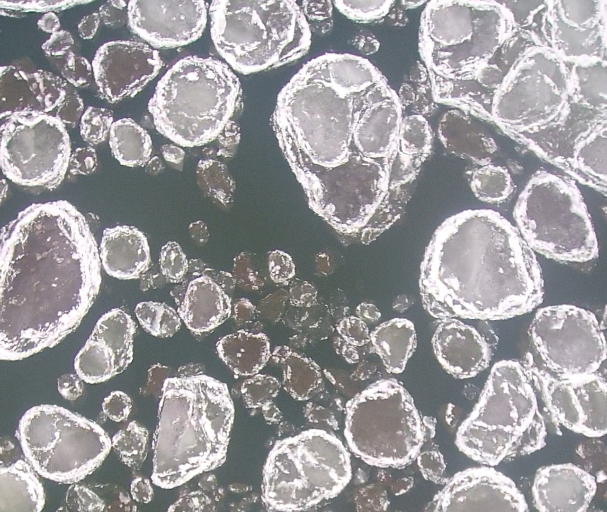}
			\includegraphics[width=0.48\linewidth]{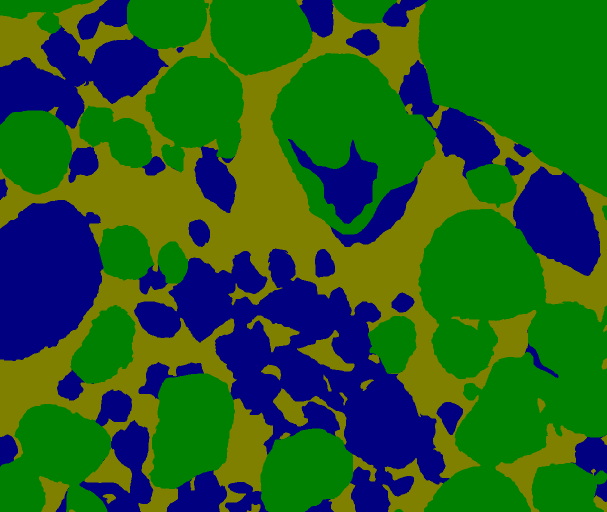}
		\end{minipage}
	}
	\subfigure[(b) Data augmentation]{
		\begin{minipage}[b]{0.60\linewidth}
			\includegraphics[width=0.22\linewidth]{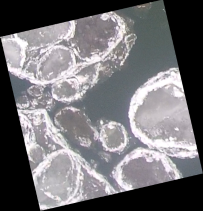}
			\includegraphics[width=0.22\linewidth]{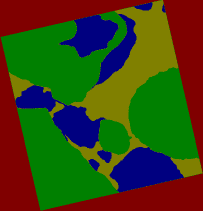}
			\includegraphics[width=0.22\linewidth]{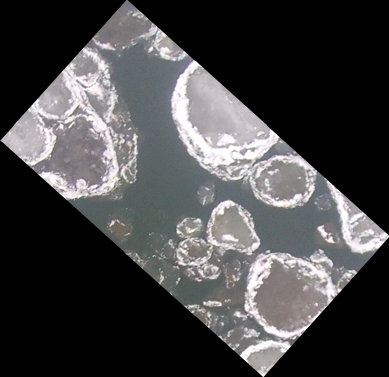}
			\includegraphics[width=0.22\linewidth]{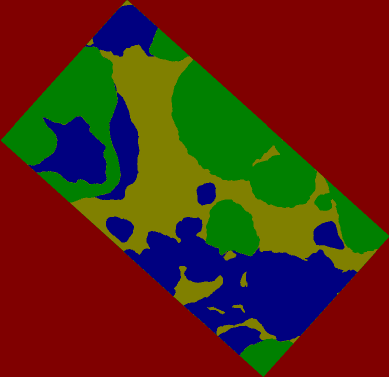}
		\end{minipage}
	}
	\caption{UAV Imagery from Alberta River Ice Segmentation Dataset. (a) the original images and labels in the dataset. (b) the images and labels processed by data augmentation.}\label{Figure7}
\end{figure}

\textbf{UAV Imagery}. There are only 50 labeled samples in the dataset, we use conventional data enhancement methods to expand them. Through random scaling, random flipping, random cropping, and random rotation, each sample was expanded 7 times, resulting in 400 samples, as shown in \textbf{Figure \ref{Figure7}}.

\begin{figure}[h]
	\centering
	\subfigure[(a)Image]{\includegraphics[width=0.22\linewidth]{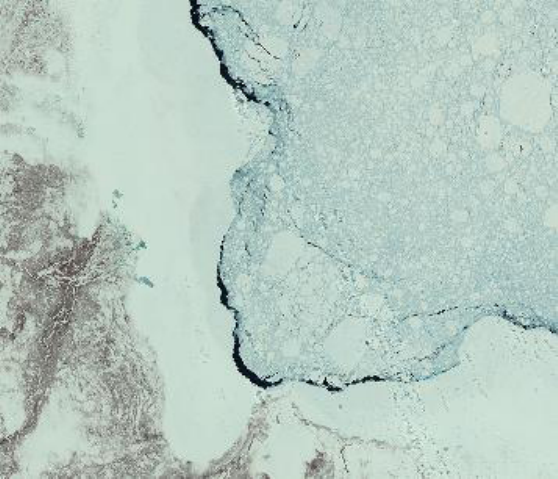}}
	\subfigure[(b)Label]{\includegraphics[width=0.22\linewidth]{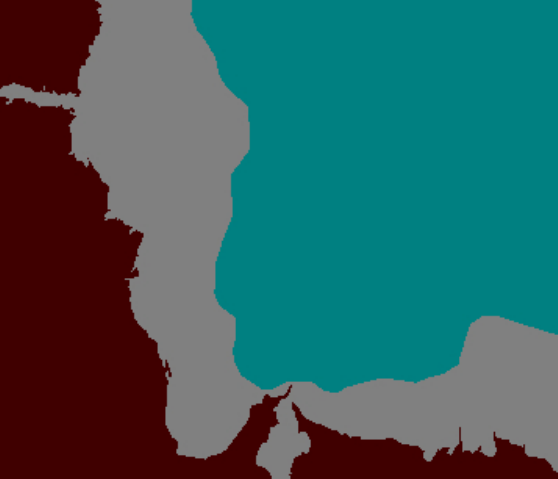}}
	\subfigure[(a)Image]{\includegraphics[width=0.22\linewidth]{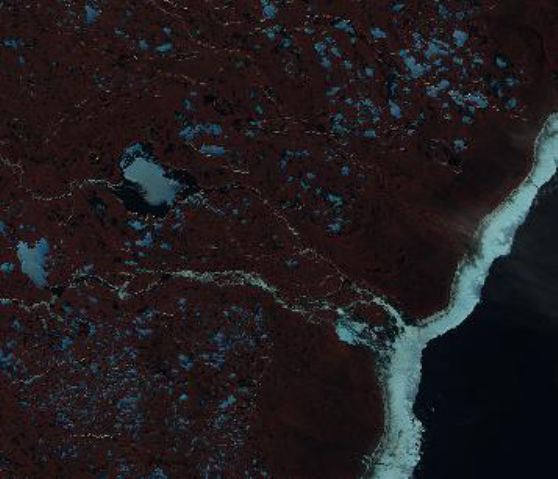}}
	\subfigure[(b)Label]{\includegraphics[width=0.22\linewidth]{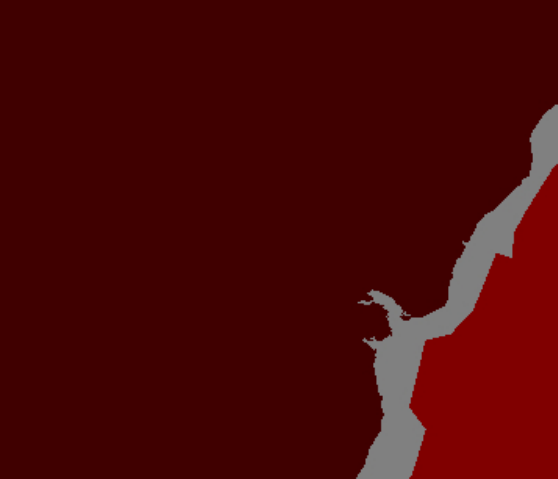}}
	\caption{Satellite Imagery from Canadian Ice Service. (a) the original images. (b) the labels for the original image.}\label{Figure8}
\end{figure}

\textbf{Satellite Imagery}. There are 3392 samples in this dataset, and the labels are in SIGRID-3 format. We convert the labels into 8 categories: water(background), 30\%ice, 50\%ice, 70\%ice, 90\%ice, 100\%ice, fastened ice, and land by existing method \cite{Neural_Net_Mapping_of_Hudson_Bay_Sea_Ice_2021}. Although this is the satellite imagery of sea ice, it is very similar to the satellite imagery of river ice. The dataset is shown in \textbf{Figure \ref{Figure8}}.

\begin{figure}[h]
	\centering
	\subfigure[(a)Image]{\includegraphics[width=0.22\linewidth]{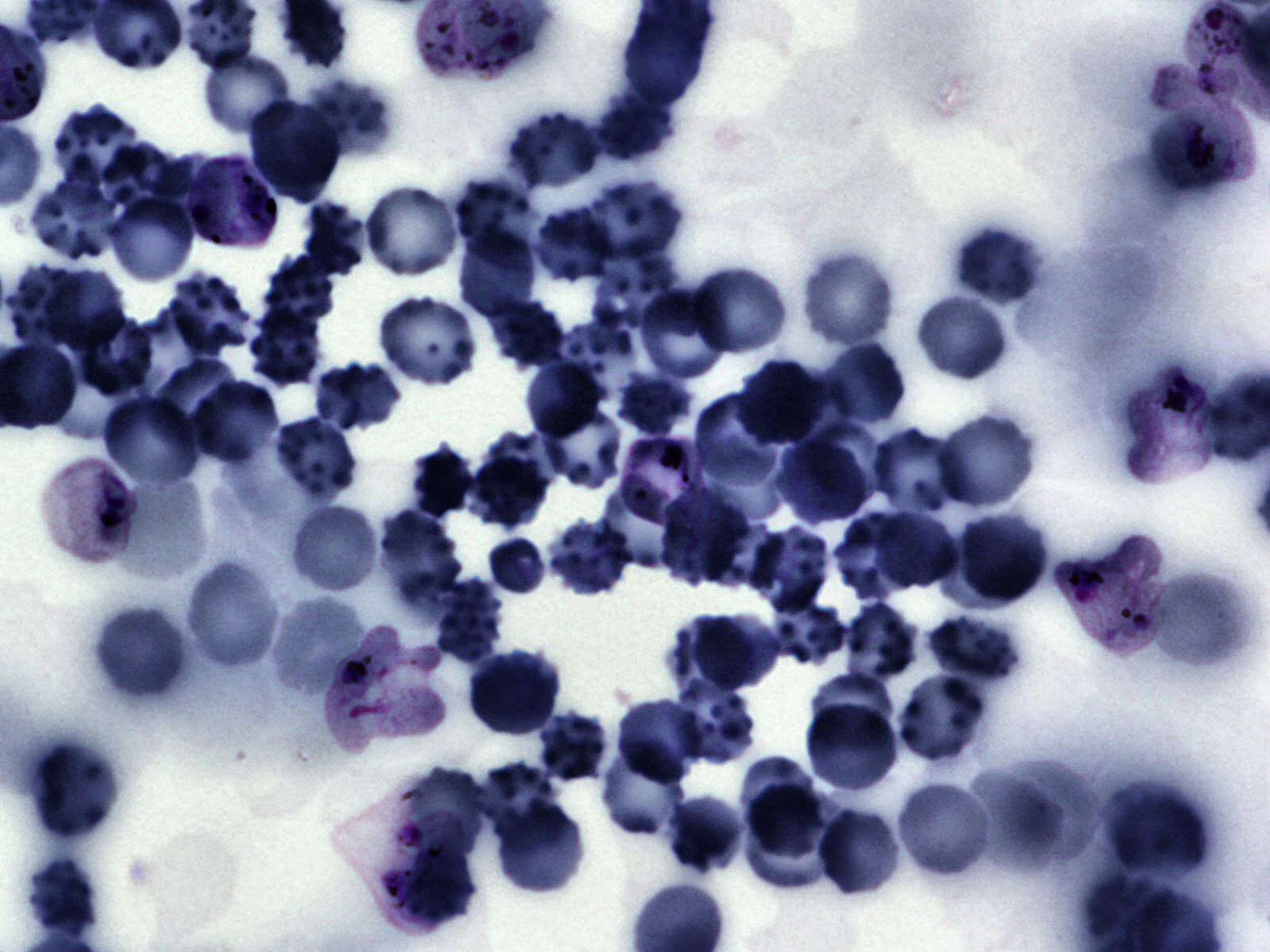}}
	\subfigure[(b)Label]{\includegraphics[width=0.22\linewidth]{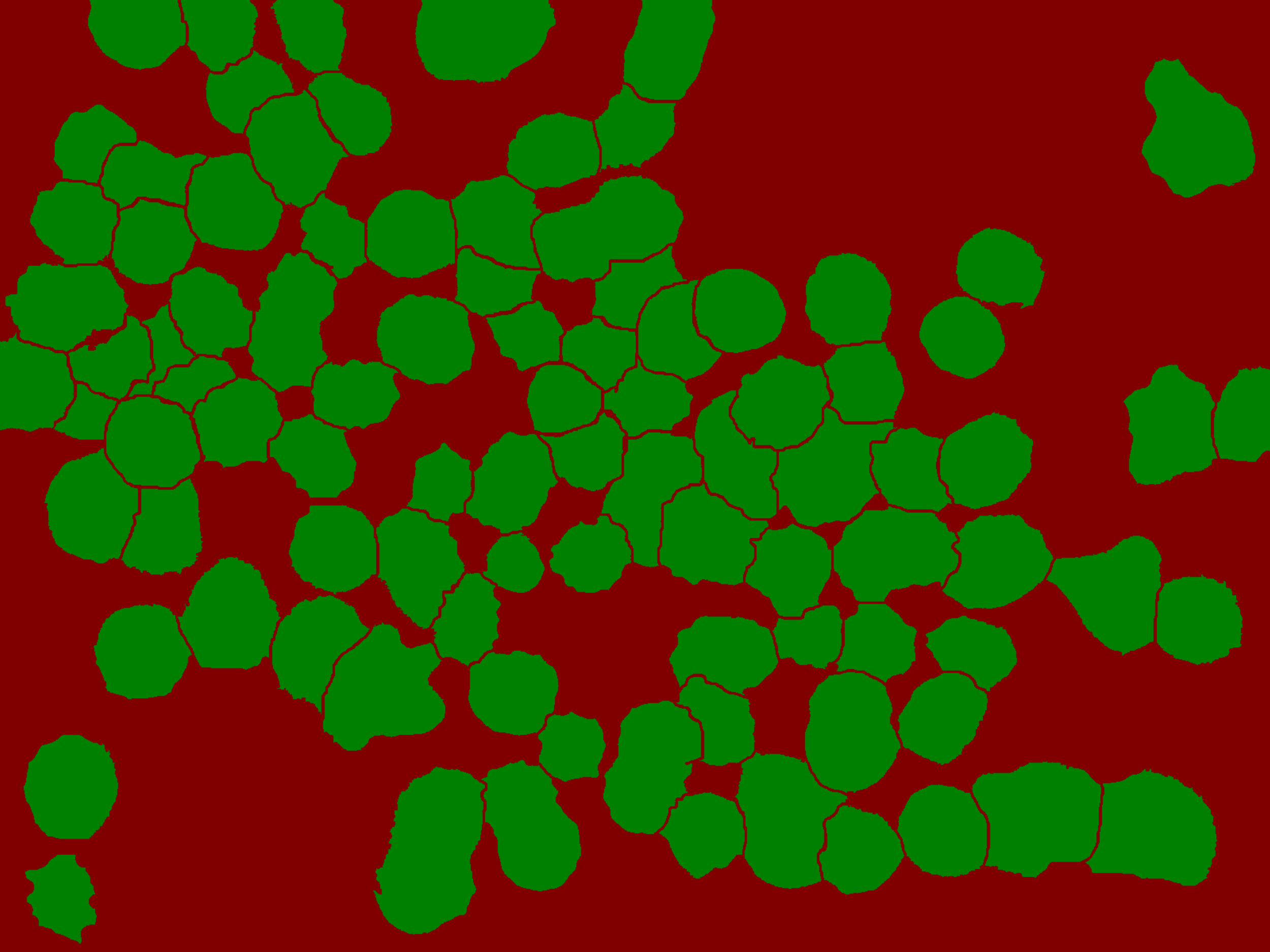}}
	\subfigure[(a)Image]{\includegraphics[width=0.22\linewidth]{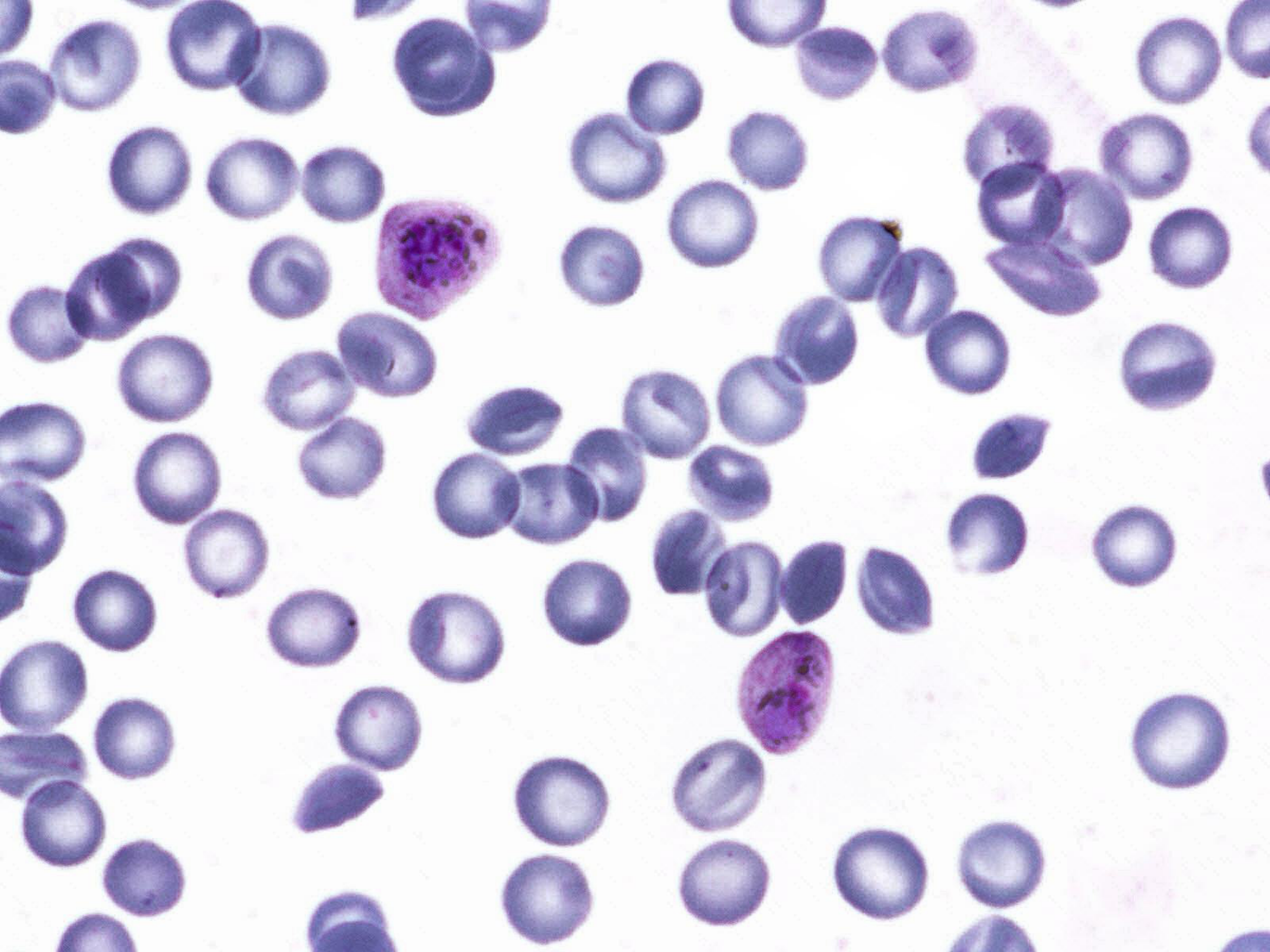}}
	\subfigure[(b)Label]{\includegraphics[width=0.22\linewidth]{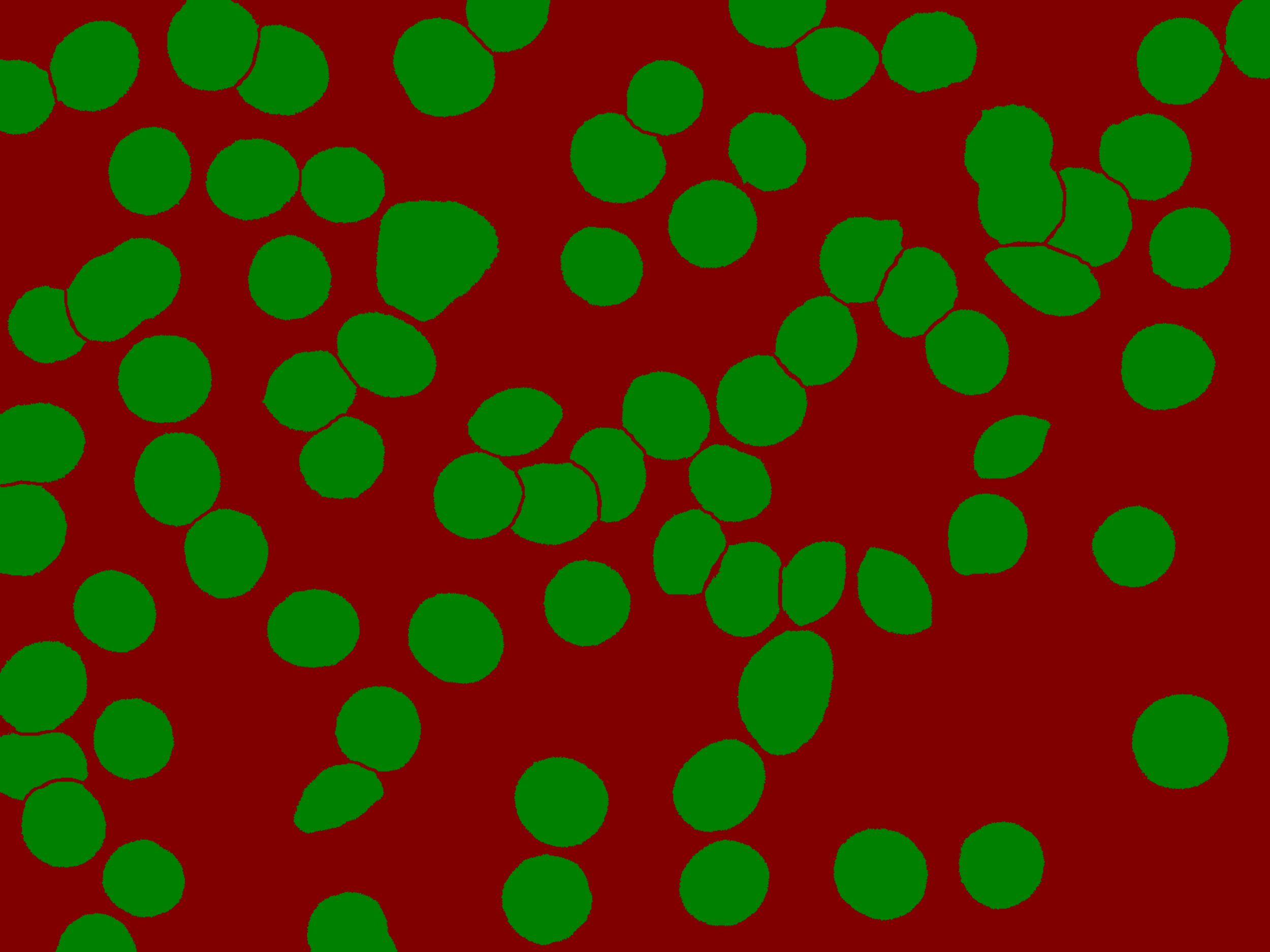}}
	\caption{Blood cell segmentation dataset. (a) the original images. (b) the labels for the original image.}\label{Figure9}
\end{figure}

\textbf{Blood Cell Segmentation Dataset}. To analyze whether medical imaging data with similar texture shapes to river ice is helpful for transfer learning of semantic segmentation of river ice, we also found a type of microscopic blood cell segmentation dataset \cite{Blood_Cell_Segmentation_Dataset_2023}, from the public datasets and used it as the training set for transfer learning. This dataset contains 1328 samples with a shape similar to the UAV Imagery mentioned above. The dataset is shown in \textbf{Figure \ref{Figure9}}.

\subsection{Evaluation Indicators}

We choose typical evaluation metrics in semantic segmentation tasks \cite{Long_Shelhamer_Darrell_2015}:
\begin{itemize}
	\item Accuracy (\textbf{Acc}): It refers to the proportion of pixels whose category prediction is correct to the total pixels. The higher the accuracy, the better the model quality;
	\item Mean intersection over the union (\textbf{mIoU}): The inference calculation is performed separately for each category dataset, the calculated intersection of the predicted area and the actual area is divided by the union of the predicted area and the actual area, and then the results obtained for all categories are averaged. The higher the average intersection ratio, the better the model quality.
\end{itemize}

There are $n+1$ categories in the dataset, and the categories are represented from $0$ to $n$, where $0$ is the background or invalid category. Let $0 \le i,j \le n$, $p_{ij}$ means that the pixel of the i-th category is predicted to be the j-th category, $p_{ii}$ means that the pixel of the i-th category is predicted to be the i-th category, and $p_{ii}$ is It's a true positive. When $i \neq j$, $p_{ij}$ and $p_{ji}$ represent false positives and false negatives.

The equation of the accuracy (Acc) is as follows:
\begin{equation}
	Acc = \frac{ {\textstyle \sum_{i=0}^{n}p_{ii}} }{ {\textstyle \sum_{i=0}^{n}} {\textstyle \sum_{j=0}^{n}}p_{ij}  } 
	\label{equation1}
\end{equation}

The equation of the Mean intersection over the union (mIoU) is as follows:
\begin{equation}
	mIoU = \frac{1}{n+1} {\textstyle \sum_{i=0}^{n}}\frac{p_{ii}}{ {\textstyle \sum_{j=0}^{n}p_{ij}}+{\textstyle \sum_{j=0}^{n}p_{ji}}-p_{ii}}  
	\label{equation2}
\end{equation}

\subsection{Semantic Segmentation of River Ice}

In this section, we first introduce the experimental implementation details of the river ice semantic segmentation network in three types of river ice datasets, secondly, introduce the comparison results with other semantic segmentation methods, and then introduce the ablation studies of our IceHrNet on the IPC\_RI\_SEG dataset.

\begin{table*}[hbt]
	\caption{Comparisons on three different river ice dataset}\label{tbl4}
	\begin{tabular*}{\linewidth}{p{18mm}p{21.5mm}*{8}{p{11.5mm}}}
		\toprule
		\multirow{3}{*}{Method} & \multirow{3}{*}{Backbone} & \multicolumn{4}{c}{Fixed Camera Imagery} & \multicolumn{2}{{c}}{\multirow{2}{*}{UAV Imagery}}  & \multicolumn{2}{{c}}{\multirow{2}{*}{Satellite Imagery}}  \\ \cline{3-6}
		& & \multicolumn{2}{c}{3 Classes} & \multicolumn{2}{c}{2 Classes} & & & &  \\ \cline{3-10}
		& & mIoU & Acc & mIoU & Acc & mIoU & Acc & mIoU & Acc \\
		\midrule
		U-Net & None & 0.9240 & 0.9637 & 0.9510 & 0.9749 & 0.7464 & 0.8716 & 0.0608 & 0.4863 \\
		UNet3+ & None & 0.9652 & 0.9804 & 0.9548 & 0.9769 & 0.7981 & 0.9047 & 0.2736 & 0.7469 \\
		CCNet & ResNet101\_vd & 0.9445 & 0.9681 & 0.9593 & 0.9792 & 0.7665 & 0.8856 & 0.3361 & 0.7648 \\
		DeepLabV3+ & ResNet101\_vd & 0.9743 & 0.9854 & 0.9642 & 0.9818 & \textbf{0.8405} & \textbf{0.9278} & 0.4262 & 0.8540 \\
		DMNet & ResNet101\_vd & 0.9711 & 0.9835 & 0.9621 & 0.9807 & 0.8169 & 0.9143 & 0.4096 & 0.8337 \\
		PSPNet & ResNet101\_vd & 0.9523 & 0.9733 & 0.9613 & 0.9803 & 0.7370 & 0.8650 & 0.3513 & 0.7962 \\
		K-Net & ResNet101\_vd & 0.9769 & 0.9869 & 0.9670 & 0.9832 & 0.7966 & 0.9015 & 0.4916 & 0.8877 \\
		SegFormer & MixViT\_B5 & 0.9743 & 0.9854 & 0.9625 & 0.9809 & 0.8110 & 0.9108 & \textbf{0.5093} & \textbf{0.8894} \\
		\hline
		\textbf{IceHrHet(ours)} & HRNet\_W48 & \textbf{0.9780} & \textbf{0.9876} & \textbf{0.9690} & \textbf{0.9843} & 0.8301 & 0.9218 & 0.4924 & 0.8841 \\
		\bottomrule
	\end{tabular*}
\end{table*}

\begin{figure*}[hbt]
	\centering
	\subfigure[(a) Daytime scene]{
		\begin{minipage}[b]{\linewidth}
			\begin{minipage}[b]{\linewidth}
				\subfigure[Image]{\includegraphics[width=0.16\linewidth]{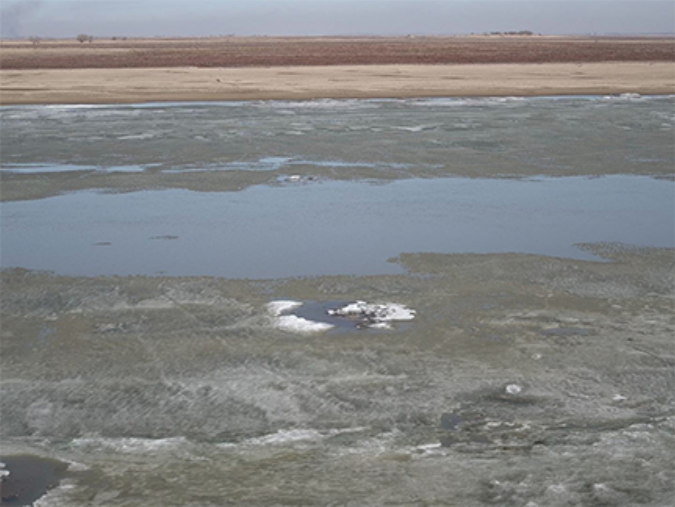}}
				\subfigure[Groundtruth]{\includegraphics[width=0.16\linewidth]{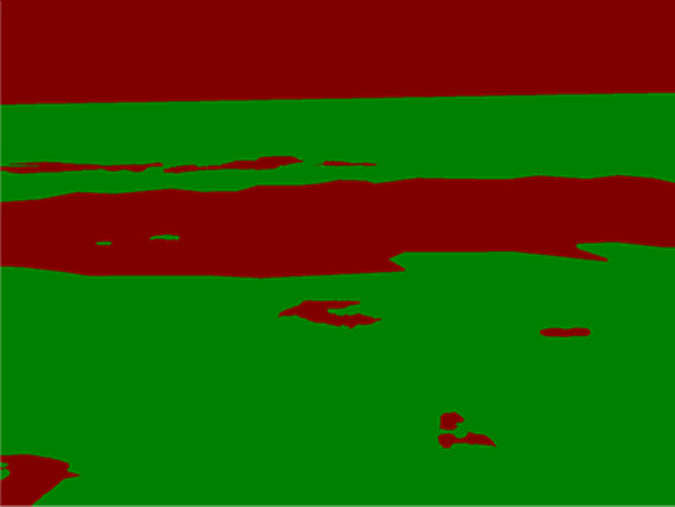}}
				\subfigure[U-Net]{\includegraphics[width=0.16\linewidth]{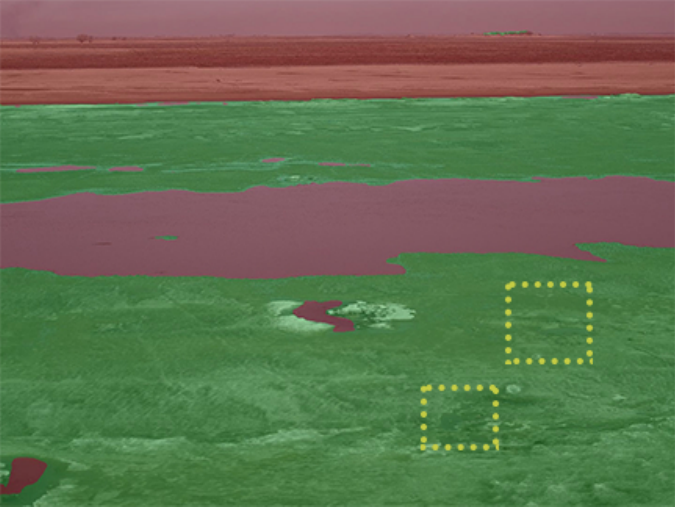}}
				\subfigure[UNet3+]{\includegraphics[width=0.16\linewidth]{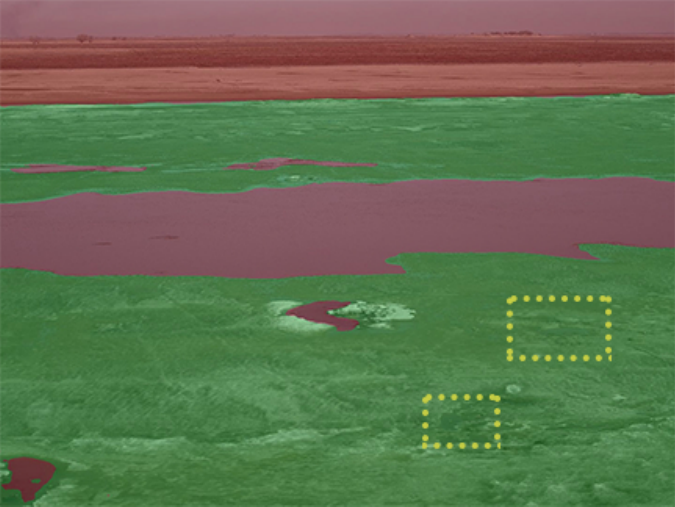}}
				\subfigure[CCNet]{\includegraphics[width=0.16\linewidth]{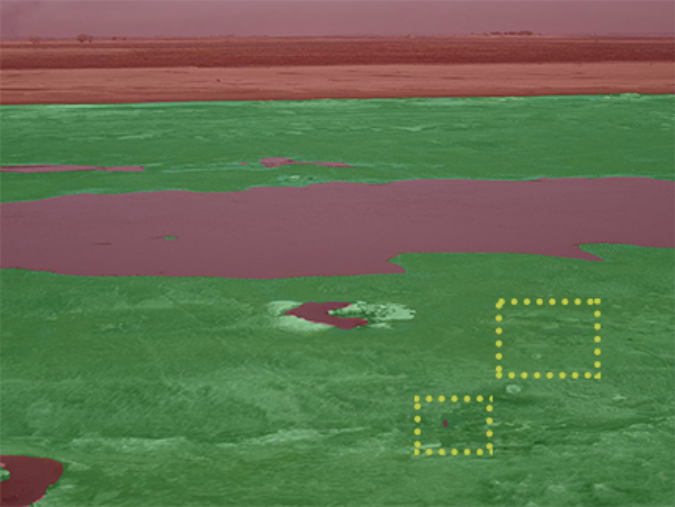}}
				\subfigure[DeepLabV3+]{\includegraphics[width=0.16\linewidth]{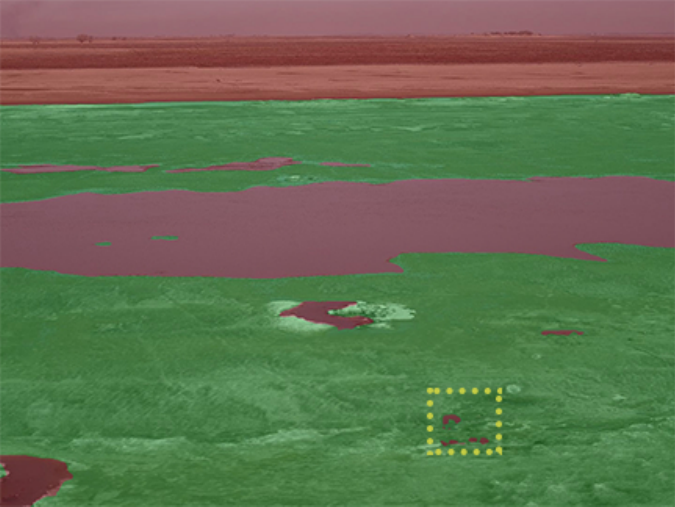}}
			\end{minipage}
			\begin{minipage}[b]{\linewidth}
				\subfigure[]{\includegraphics[width=0.16\linewidth]{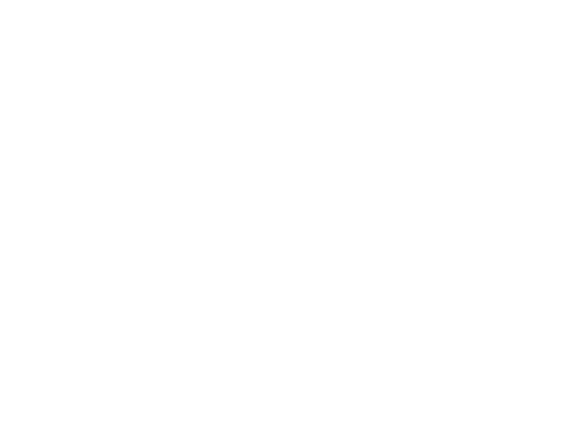}}
				\subfigure[DMNet]{\includegraphics[width=0.16\linewidth]{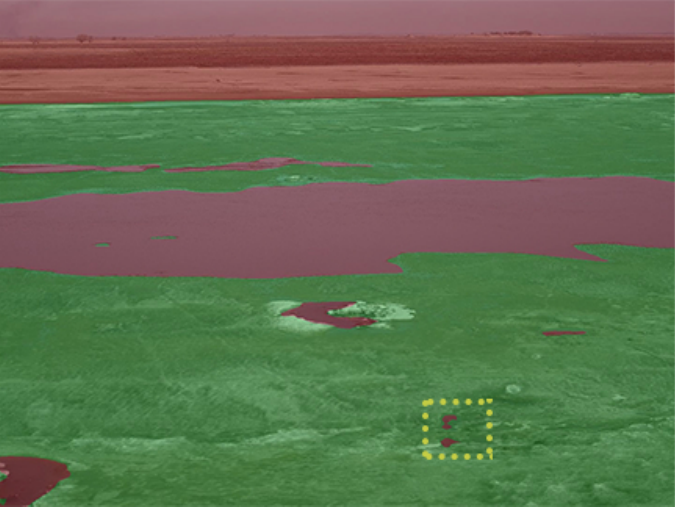}}
				\subfigure[PSPNet]{\includegraphics[width=0.16\linewidth]{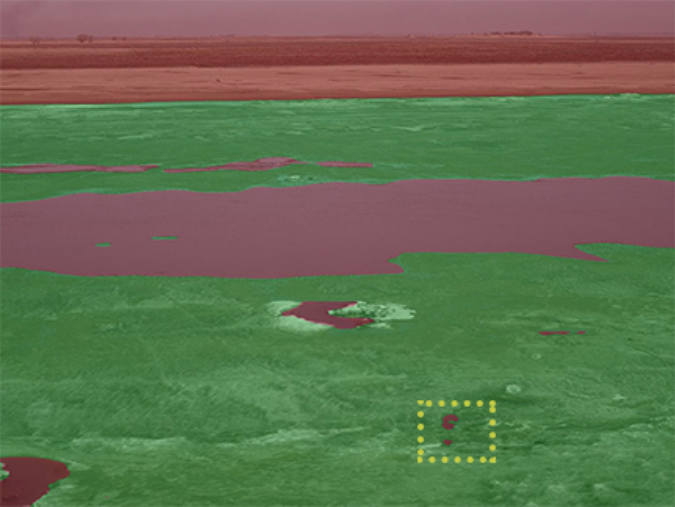}}
				\subfigure[K-Net]{\includegraphics[width=0.16\linewidth]{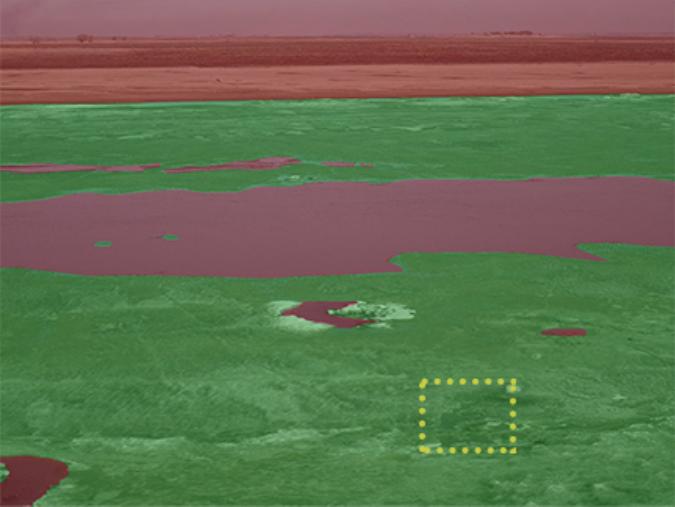}}
				\subfigure[SegFormer]{\includegraphics[width=0.16\linewidth]{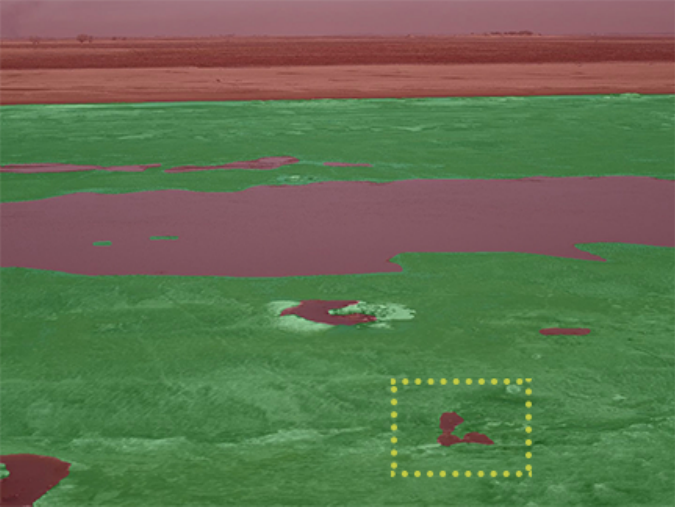}}
				\subfigure[\textbf{IceHrNet(ours)}]{\includegraphics[width=0.16\linewidth]{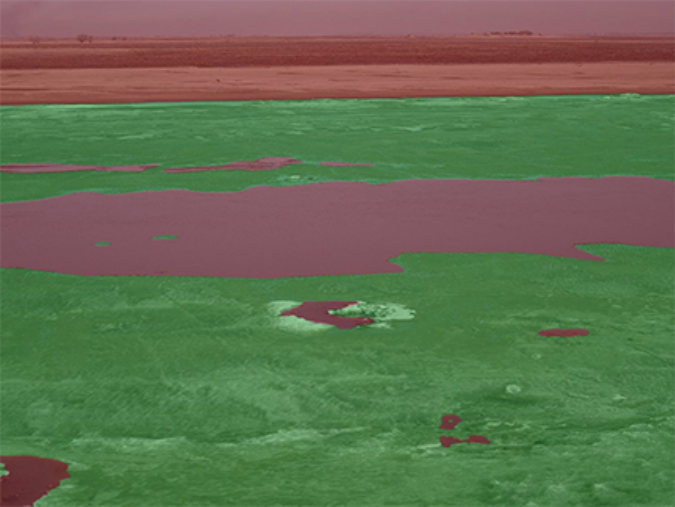}}
			\end{minipage}
		\end{minipage}
	}
	
	\subfigure[(b) Night scene]{
		\begin{minipage}[b]{\linewidth}
			\begin{minipage}[b]{\linewidth}
				\subfigure[Image]{\includegraphics[width=0.16\linewidth]{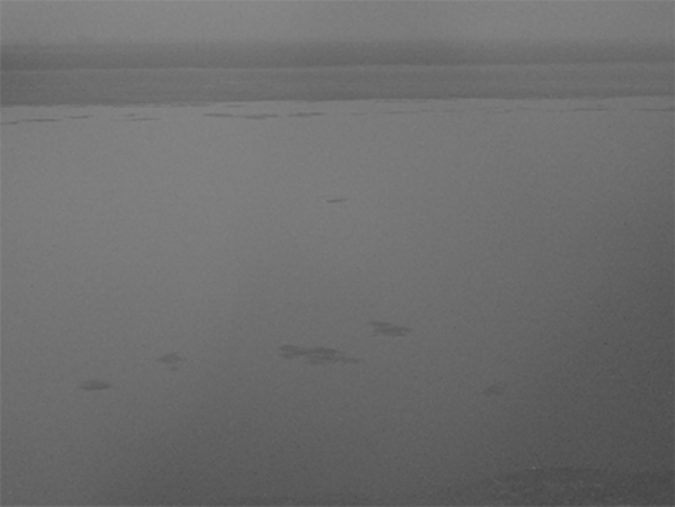}}
				\subfigure[Groundtruth]{\includegraphics[width=0.16\linewidth]{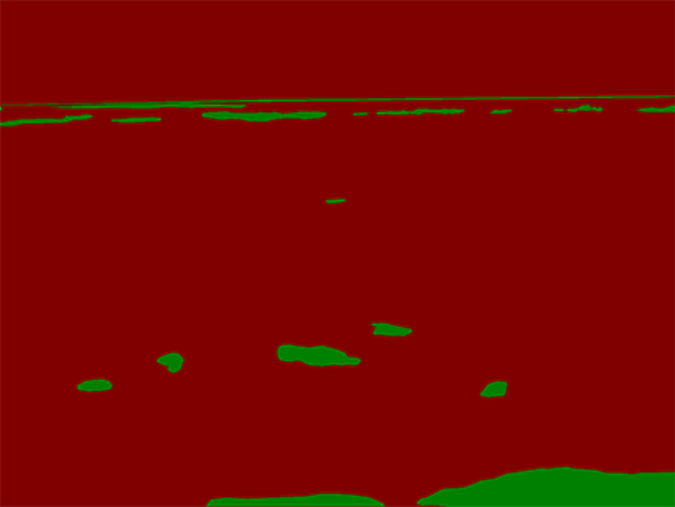}}
				\subfigure[U-Net]{\includegraphics[width=0.16\linewidth]{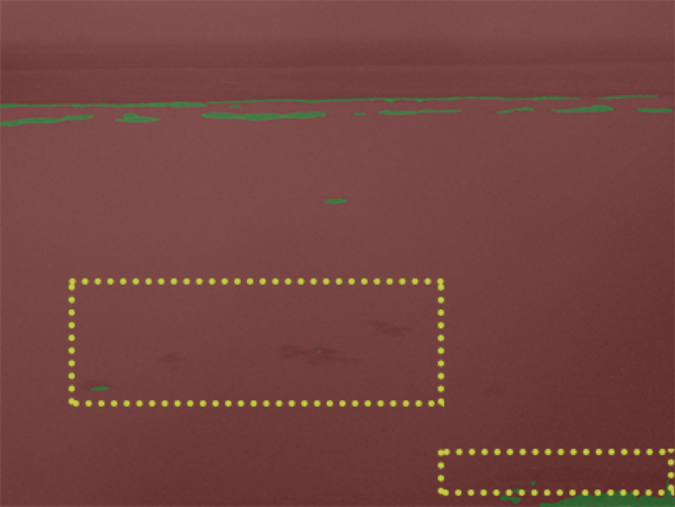}}
				\subfigure[UNet3+]{\includegraphics[width=0.16\linewidth]{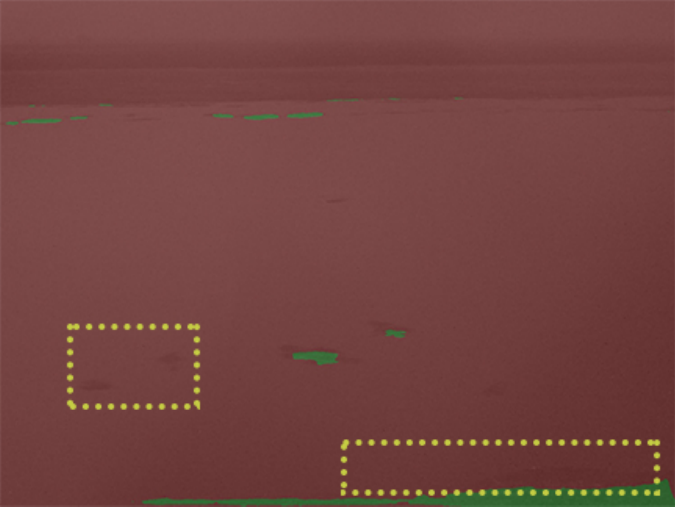}}
				\subfigure[CCNet]{\includegraphics[width=0.16\linewidth]{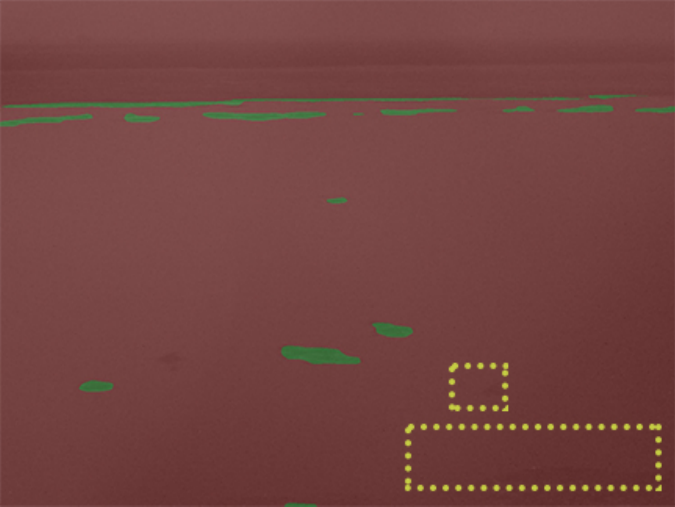}}
				\subfigure[DeepLabV3+]{\includegraphics[width=0.16\linewidth]{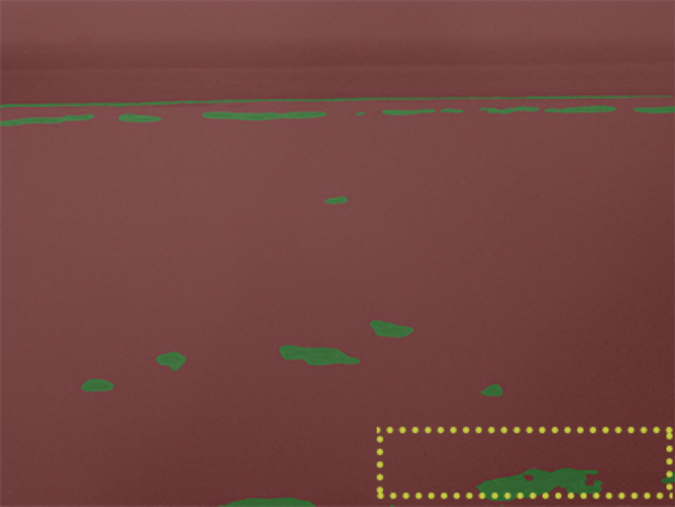}}
			\end{minipage}
			\begin{minipage}[b]{\linewidth}
				\subfigure[]{\includegraphics[width=0.16\linewidth]{blank.png}}
				\subfigure[DMNet]{\includegraphics[width=0.16\linewidth]{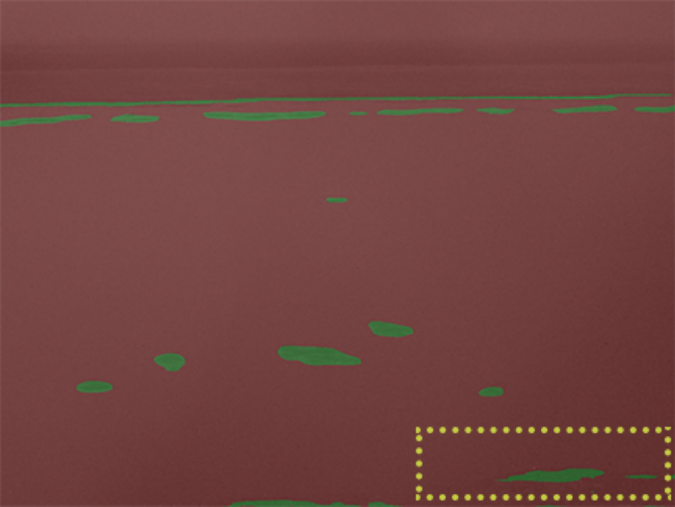}}
				\subfigure[PSPNet]{\includegraphics[width=0.16\linewidth]{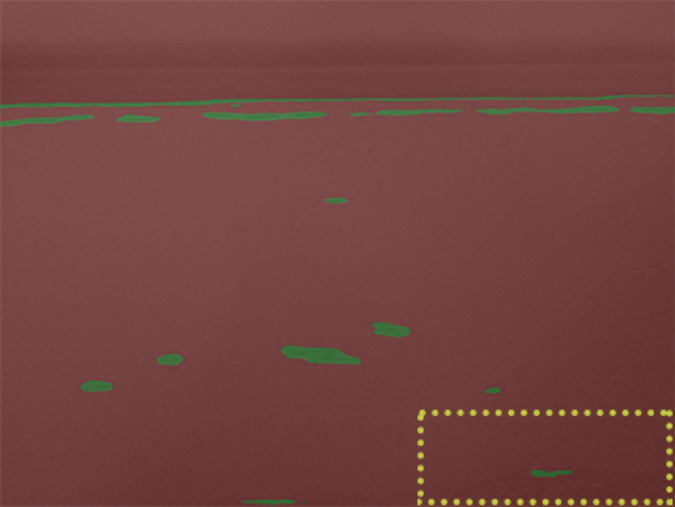}}
				\subfigure[K-Net]{\includegraphics[width=0.16\linewidth]{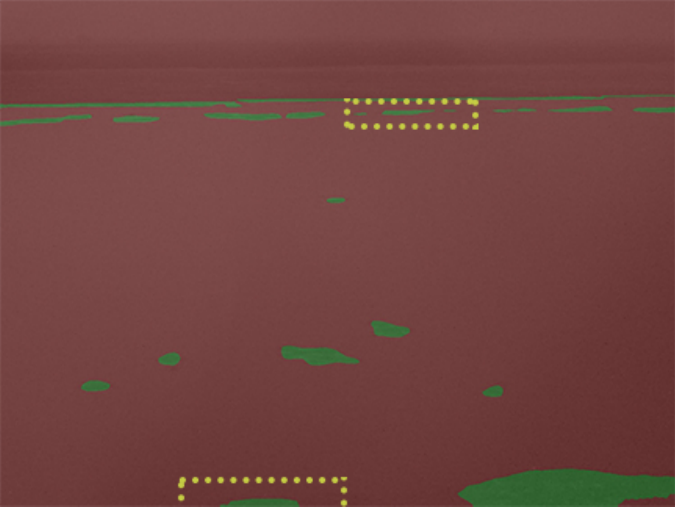}}
				\subfigure[SegFormer]{\includegraphics[width=0.16\linewidth]{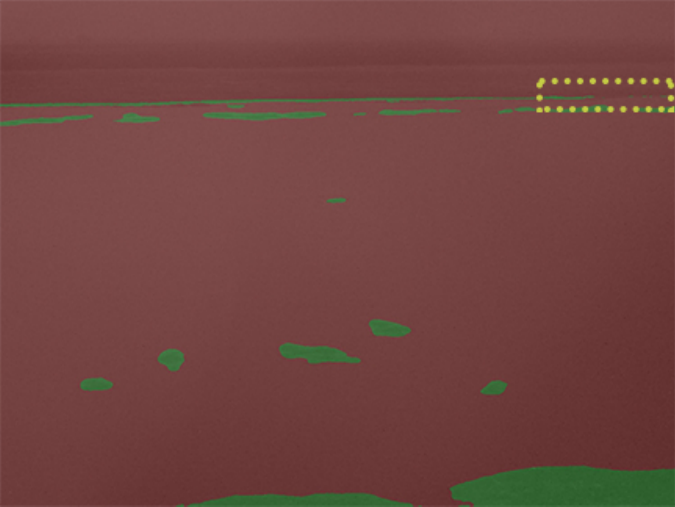}}
				\subfigure[\textbf{IceHrNet(ours)}]{\includegraphics[width=0.16\linewidth]{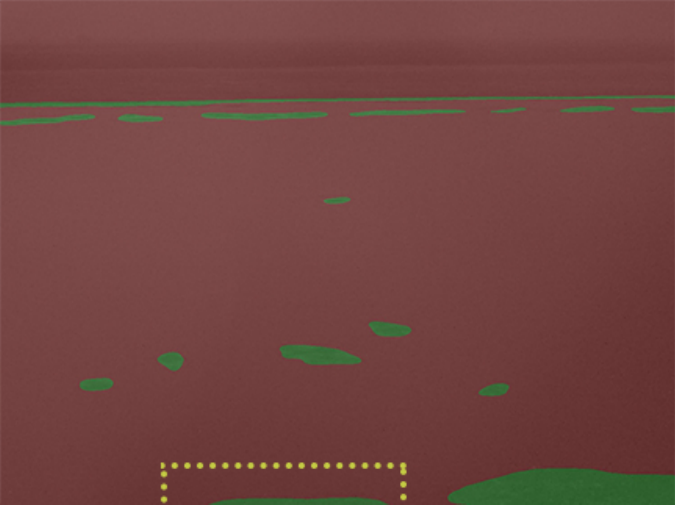}}
			\end{minipage}
		\end{minipage}
	}
	\caption{Visualization results on the test set of 2-classes Fixed Camera Imagery. CNN-based model errors mostly occur in the same place, the ViT-based model SegFomer errors occurred elsewhere. Although IceHrNet is not significantly different from others overall, it emphasizes the recognition of texture details more. (a) Comparison of daytime scenes.(b) Comparison of night scenes.}\label{Figure10}
\end{figure*}

\subsubsection{Implementation Details}

To compare the performance of our deep convolutional neural network model IceHrNet, we chose state-of-the-art methods in recent years for comparison. The code of the state-of-the-art methods used in the paper was implemented in the PaddleSeg project \cite{liu2021paddleseg}. They are CCNet \cite{Huang_Wang_Huang_Huang_Wei_Liu_2019}, DeepLabV3+ \cite{Chen_Zhu_Papandreou_Schroff_Adam_2018}, DM Net \cite{He_Deng_Qiao_2019}, PSPNet \cite{Zhao_Shi_Qi_Wang_Jia_2017}, SegFormer \cite{Xie_Wang_Yu_Anandkumar_Alvarez_Luo_2021}, UNet3+ \cite{Huang_Lin_Tong_Hu_Zhang_Iwamoto_Han_Chen_Wu_2020}, U-Net \cite{Ronneberger_Fischer_Brox_2015}, K-Net \cite{Zhang_Pang_Chen_Loy_2021}. Most of them are based on convolutional neural network methods, except SegFormer which is based on ViT \cite{Dosovitskiy_Beyer_Kolesnikov_Weissenborn_Zhai_2020}. U-Net and UNet3+, are effective in the field of medical image segmentation. We used these methods to conduct experimental comparisons on three types of river ice datasets. The training set, validation set, and test set are uniformly distributed in a ratio of 6:2:2.

In the training stage, we used the AdmW optimizer and the learning rate strategy adopted MultiStepDecay, as shown in \textbf{Table \ref{tbl3}}. The cross-entropy loss function is used, and the number of training iterations is set to 40,000.

\begin{table}[h]
	\centering
	\caption{Settings of optimizer add lr\_scheduler}\label{tbl3}
	\begin{tabular*}{\tblwidth}{@{}ll@{}}
		\toprule
		\textbf{AdamW} & \\
		\midrule
		weight\_decay & 0.005 \\
		grad\_clip\_cfg.name & ClipGradByNorm \\
		grad\_clip\_cfg.clip\_norm & 1 \\
		\toprule[0.5pt]
		\textbf{MultiStepDecay} & \\
		\midrule
		milestones & [30000,36000] \\
		warmup\_iters & 1000 \\
		warmup\_start\_lr & 0.00001 \\
		learning\_rate & 0.0001 \\
		\bottomrule
	\end{tabular*}
\end{table}

\subsubsection{Experimental Results}

In the experiment, nine different methods were trained and tested on three types of river ice datasets. The number of categories of Fixed Camera Imagery in the dataset is 3 and 2, the number of categories in UAV Imagery is 4, and the number of categories in Satellite Imagery is 8. We plot the experimental results on a table, see \textbf{Table \ref{tbl4}}.

From the experimental results, it can be seen that the proposed IceHrNet outperforms the other methods on the texture-focused dataset Fixed Camera Imagery. Although the improvement is not much, it will be a great help for accurate river ice density calculation. The visual result comparison is shown in \textbf{Figure \ref{Figure10}}. This feature could effectively support our subsequent zero-shot transfer learning based on style transferring. 

In the UAV Imagery dataset, which focuses on shape and texture features, DeepLabV3+ achieved the best performance. Its deep residual network is indeed very convenient in extracting advanced semantics of shapes. However, IceHrNet also achieved a good score in second place.

In the Satellite Imagery dataset, which mainly focuses on texture features, the overall score decreased due to sample imbalance. The SegFormer pretrained on a large-scale dataset based on ViT demonstrated its advantages and achieved the best score, while IceHrNet also performed well and achieved the second highest score, which was 6 percentage points higher than DeepLabV3+, better demonstrating the effectiveness of IceHrNet in segmenting texture-focused datasets.

\subsubsection{Ablation Study}

For the following ablation experiments, we utilize UAV Imagery which has 4 categories as the dataset. Fixed random seed during the experiment. Remove the ASPP unit and the Decoder unit respectively and replace the low-level feature maps Conv1 and Conv2 used in the Decoder, to observe the impact of different stages on the model. The baseline is the HRNet backbone, which adds a sample FCNHead for segmentation. See \textbf{Table \ref{tbl5}}.

\begin{table}[htbp]
	\centering
	\caption{Ablation study of exchange units. (a) is the baseline using a sample FCNHead to classify pixels. (e) is the final method of IceHrNet.}\label{tbl5}
	\begin{tabular*}{\tblwidth}{p{8mm}p{11mm}p{8mm}*{4}{p{6.5mm}}}
		\toprule
		Method & FCNHead & Decoder & ASPP & Conv1 & Conv2 & mIoU \\
		\midrule
		(a) & \checkmark & & & & & 0.8197  \\
		(b) &  & \checkmark & & & \checkmark & 0.7894  \\
		(c) &  & \checkmark & \checkmark & & \checkmark & 0.7985  \\
		(d) &  & \checkmark & & \checkmark & & 0.8114  \\
		(e) &  & \checkmark & \checkmark & \checkmark & & \textbf{0.8301}  \\
		\bottomrule
	\end{tabular*}
\end{table}

Method (a), HRNet is used as the baseline, which removes the ASPP unit and the Decoder unit from IceHrNet, and then uses a simple FCNHead as the segmentation heads. The FCNHead is composed of two convolutional layers. The first layer is a ConvBNReLu with kernel=1, stride=1, and out\_channels unchanged. The second layer is a Conv2D with kernel=1, stride=1, and out\_channels=4(classes). Finally, the prediction The output is bilinearly scaled to the original image size. The FCNHead method can be used as a baseline for ablation experiments. 

Method (b), based on Method (a), removes the FCNHead, adds the Decoder unit, and uses the Conv2 for a low-level feature map. Observe the impact of the Decoder on network performance when using the Conv2 feature map. The network performance has not been improved.

Method (c), based on Method (b), adds the ASPP unit and uses the Conv2 for a low-level feature map. Observe the impact of the ASPP on network performance. The network effect is slightly improved than that without the ASPP unit.

Method (d), based on Method (a), removes the FCNHead, adds the Decoder unit, and uses the Conv1 for a low-level feature map. Observe the impact of the Decoder on network performance when using the Conv1 feature map. The network using Conv1 is much better than the network using Conv2.

Method (e), based on Method (d), adds the ASPP unit and uses the Conv1 for a low-level feature map. Observe the impact of the ASPP on network performance. The network effect using the ASPP unit and Conv1 features has been greatly improved.

\subsection{Style Transfer Learning}

In this section, we introduce the experimental results of our proposed advanced style transfer learning strategy, which is the main innovation of this paper and makes the work complete.
\begin{table*}[htbp]
	\centering
	\caption{Style Transfer Learning experiments on Fixed Camera Imagery dataset.}\label{tbl6}
	\begin{tabular*}{\tblwidth}{*{2}{p{24mm}<{\centering}}*{3}{p{34.7mm}<{\centering}}}
		\toprule
		\multirow{3}{*}{Method} & \multicolumn{4}{c}{Fixed Camera Imagery} \\
		\cline{2-5}
		& \multirow{2}{*}{Supervised} & \multicolumn{3}{c}{Blood Cell Imagery} \\
		\cline{3-5}
		&  & None Stylized & Conventional Stylized & Advanced Stylized \\
		\midrule
		DeepLabV3+ & 0.9642 & \textbf{0.2679} & 0.3213 & 0.8286 \\
		K-Net & 0.9670 & 0.2527 & 0.3515 & 0.8351 \\
		IceHrNet(ours) & 0.9690 & 0.2522 & \textbf{0.6239} & \textbf{0.8512} \\
		\bottomrule
	\end{tabular*}
\end{table*}

\begin{table*}[htbp]
	\centering
	\caption{Style Transfer Learning Experiments on UAV Imagery dataset.}\label{tbl7}
	\begin{tabular*}{\tblwidth}{*{2}{p{24mm}<{\centering}}*{3}{p{34.7mm}<{\centering}}}
		\toprule
		\multirow{3}{*}{Method} & \multicolumn{4}{c}{UAV Imagery} \\
		\cline{2-5}
		& \multirow{2}{*}{Supervised} & \multicolumn{3}{c}{Blood Cell Imagery} \\
		\cline{3-5}
		&  & None Stylized & Conventional Stylized & Advanced Stylized \\
		\midrule
		DeepLabV3+ & 0.9422 & 0.2773 & 0.3740 & 0.8645 \\
		K-Net & 0.9254 & \textbf{0.4282} & \textbf{0.4371} & 0.7520 \\
		IceHrNet(ours) & 0.9421 & 0.3365 & 0.3259 & \textbf{0.8704} \\
		\bottomrule
	\end{tabular*}
\end{table*}

\subsubsection{Implementation Details}

The experiment has a blood cell dataset in the medical imaging domain with semantic segmentation labels, and a small amount or an unlabeled river ice image in the target domain. The goal is to use the existing medical imaging datasets to train any semantic segmentation network, and then directly apply the trained model to the target domain, to observe whether the segmentation results are effective. The experiments were conducted on the 2-classes Fixed Camera Imagery and 2-classes UAV Image datasets respectively. Four sets of experiments are conducted for each dataset:
\begin{enumerate}
	\item \textbf{Supervised}. Directly use the target dataset to train the models, and the results are used as upper-limit comparisons.
	\item \textbf{None Stylized}. Only use the source dataset to train the models; the source dataset does not undergo any style transferring.
	\item \textbf{Conventional Stylized}. Use the limited target dataset images to perform the conventional global-image style transfer strategy \cite{Jackson_Atapour_Abarghouei_Bonner_Breckon_Obara_2019, Li_Ye_Cao_Hou_Yang_2021, Zhao_Wei_Lu_Bai_Zhao_Chen_Hu_2023} on the source dataset, and then train the models.
	\item \textbf{Advanced Stylized}. Use the limited target dataset images to perform our advanced style transfer learning strategy on the source dataset, and then train the models.
\end{enumerate}  

For semantic segmentation methods, we chose K-Net, DeepLabV3+, and IceHrNet, which had better segmentation results in the previous section. The neural style transfer method uses AdaIN \cite{Huang_Belongie_2017}.

\subsubsection{Experimental Comparison}

There are a total of 8 groups in the experiment, 4 groups for each dataset.

\begin{figure*}[hbt]
	\centering
	\scalebox{0.9}{
		\begin{minipage}[b]{\linewidth}
			\begin{minipage}[b]{\linewidth}
				\begin{minipage}[b]{0.24\linewidth}
					\centering
					\textbf{(a) Original Image}
				\end{minipage}
				\begin{minipage}[b]{0.24\linewidth}
					\centering
					\textbf{(b) None Stylized}
				\end{minipage}
				\begin{minipage}[b]{0.24\linewidth}
					\centering
					\textbf{(c) Conventional Stylized}
				\end{minipage}
				\begin{minipage}[b]{0.24\linewidth}
					\centering
					\textbf{(d) Advanced Stylized}
				\end{minipage}
			\end{minipage}
			\begin{minipage}[b]{\linewidth}
				\subfigure[Fixed Camera Imagery]{\includegraphics[width=0.24\linewidth]{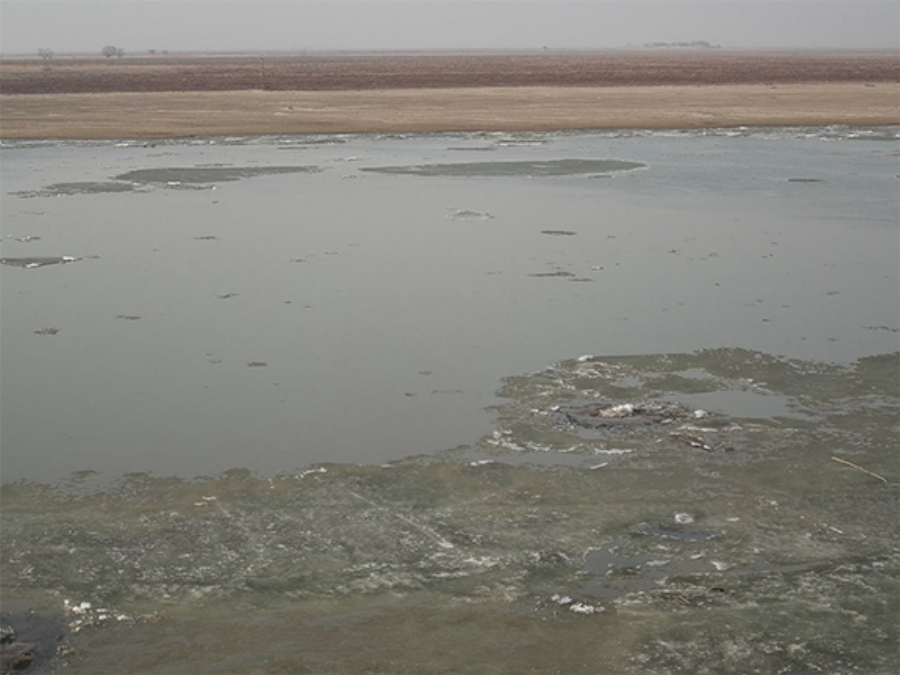}}
				\subfigure[DeepLabV3+]{\includegraphics[width=0.24\linewidth]{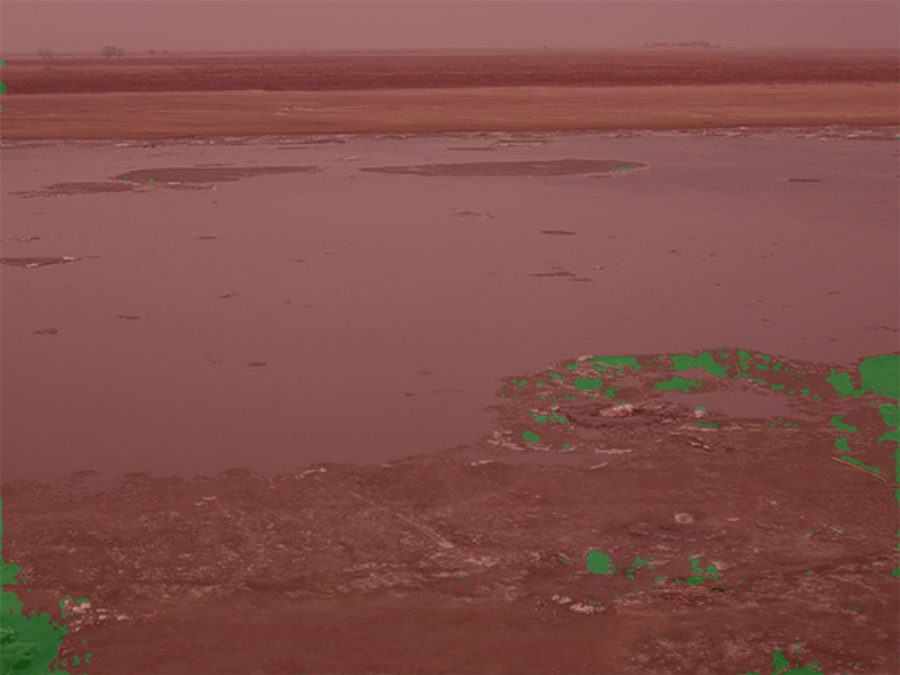}}
				\subfigure[DeepLabV3+]{\includegraphics[width=0.24\linewidth]{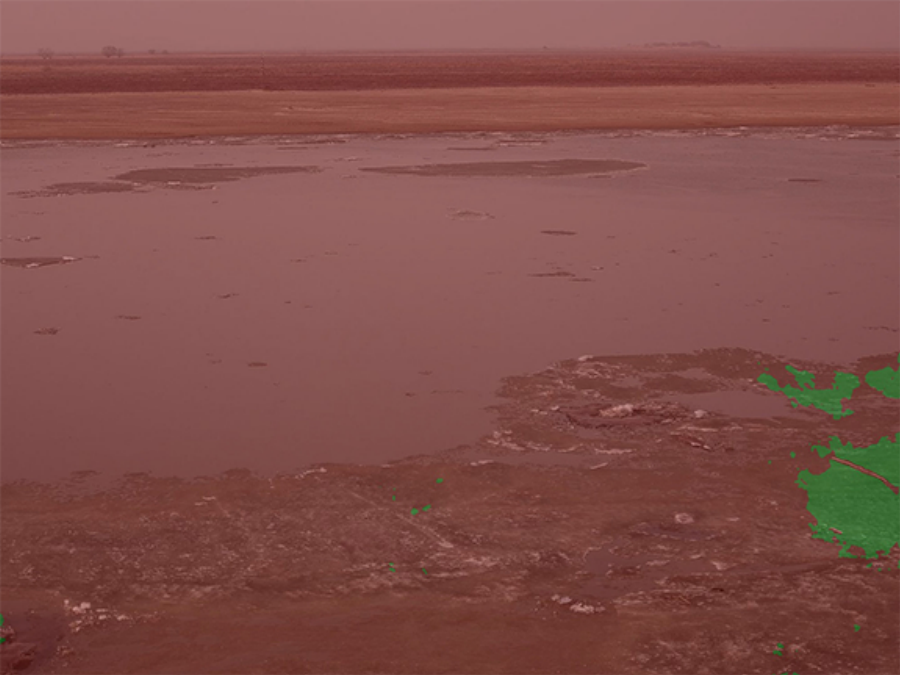}}
				\subfigure[DeepLabV3+]{\includegraphics[width=0.24\linewidth]{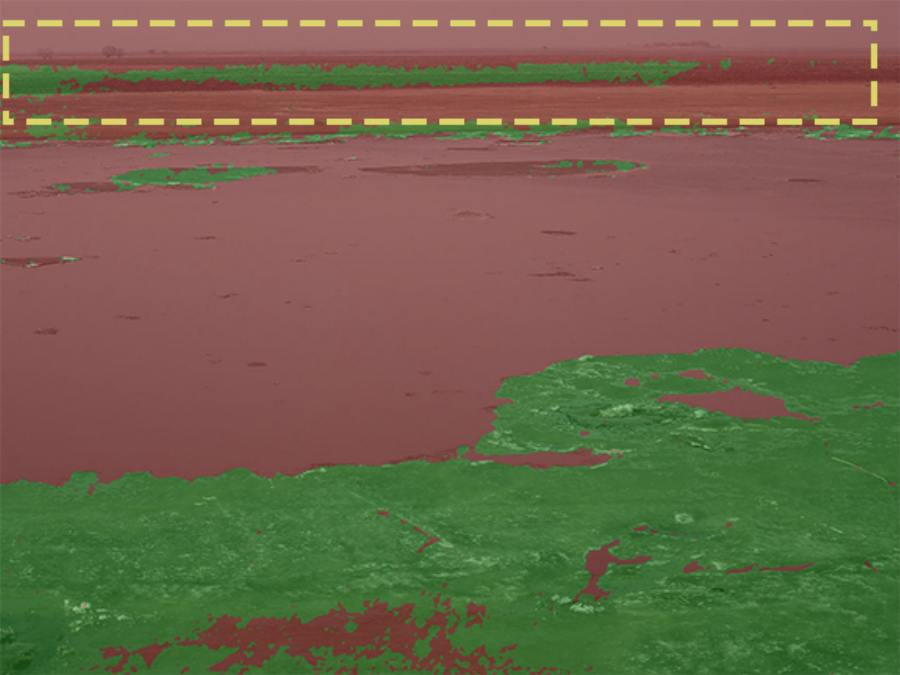}}
			\end{minipage}
			\begin{minipage}[b]{\linewidth}
				\subfigure[]{\includegraphics[width=0.24\linewidth]{blank.png}}
				\subfigure[K-Net]{\includegraphics[width=0.24\linewidth]{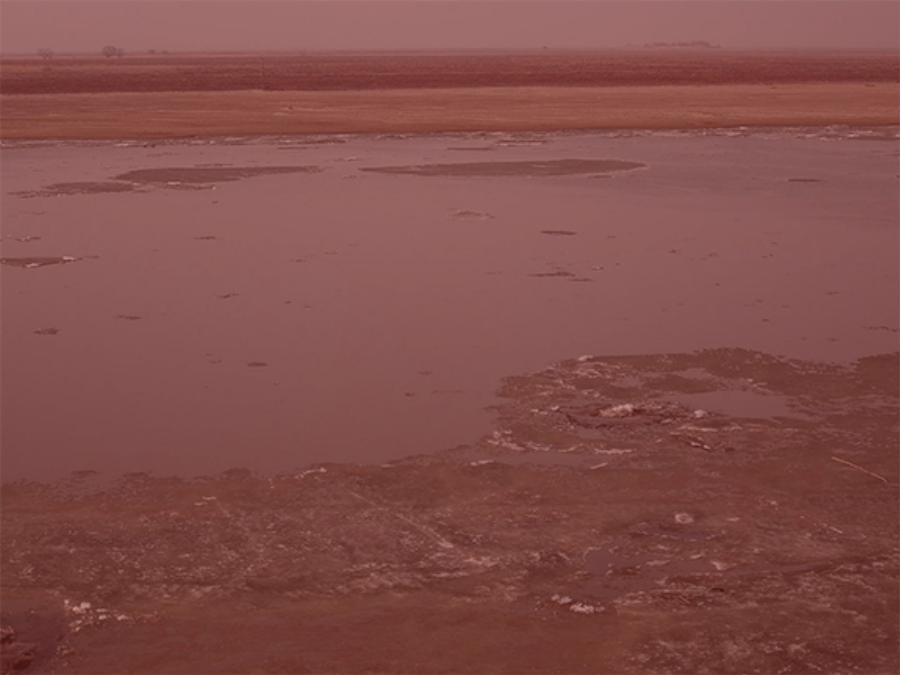}}
				\subfigure[K-Net]{\includegraphics[width=0.24\linewidth]{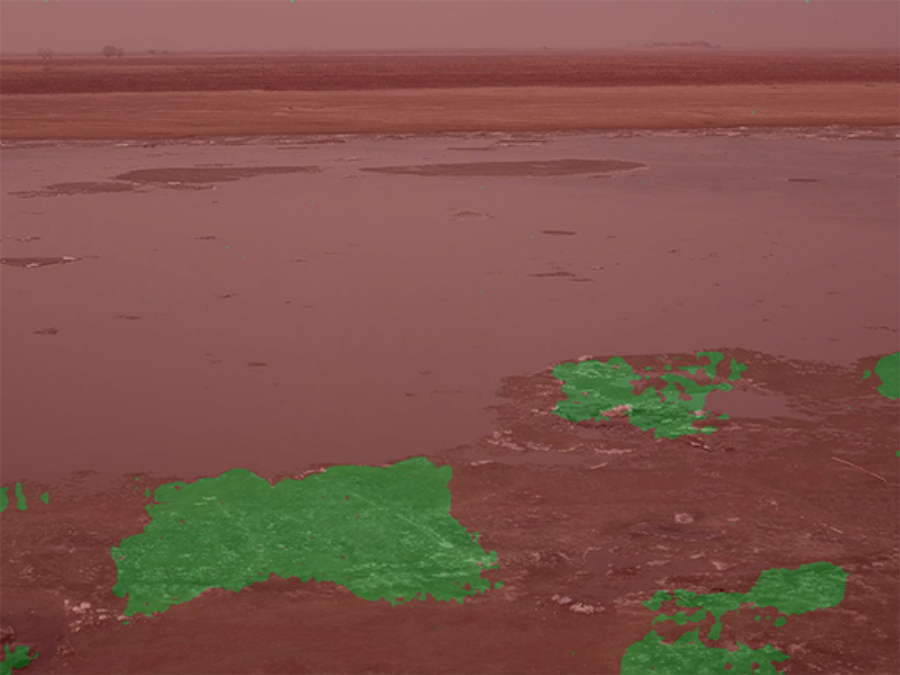}}
				\subfigure[K-Net]{\includegraphics[width=0.24\linewidth]{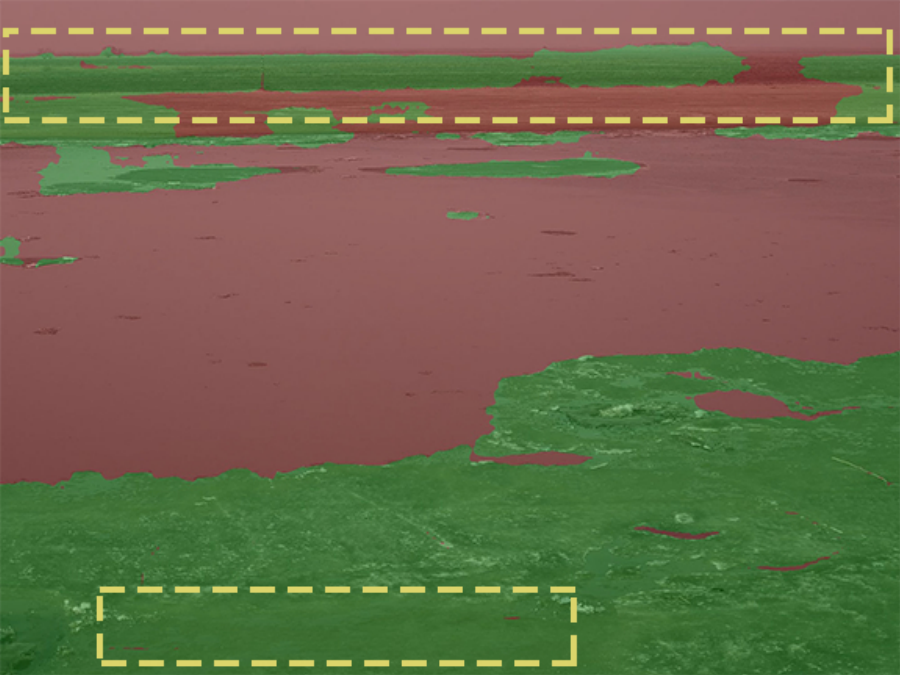}}
			\end{minipage}
			\begin{minipage}[b]{\linewidth}
				\subfigure[]{\includegraphics[width=0.24\linewidth]{blank.png}}
				\subfigure[IceHrNet(ours)]{\includegraphics[width=0.24\linewidth]{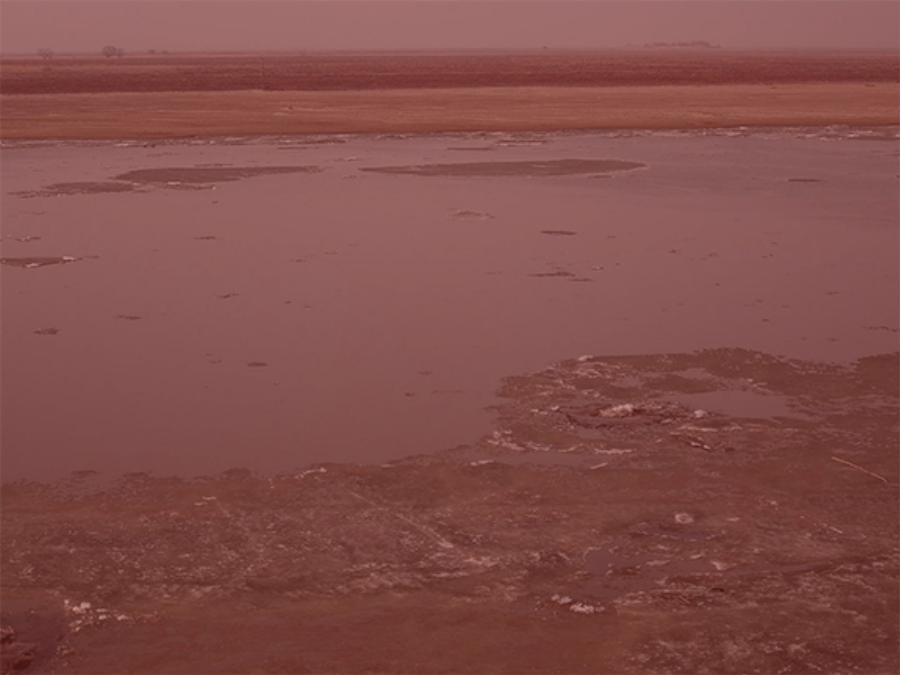}}
				\subfigure[IceHrNet(ours)]{\includegraphics[width=0.24\linewidth]{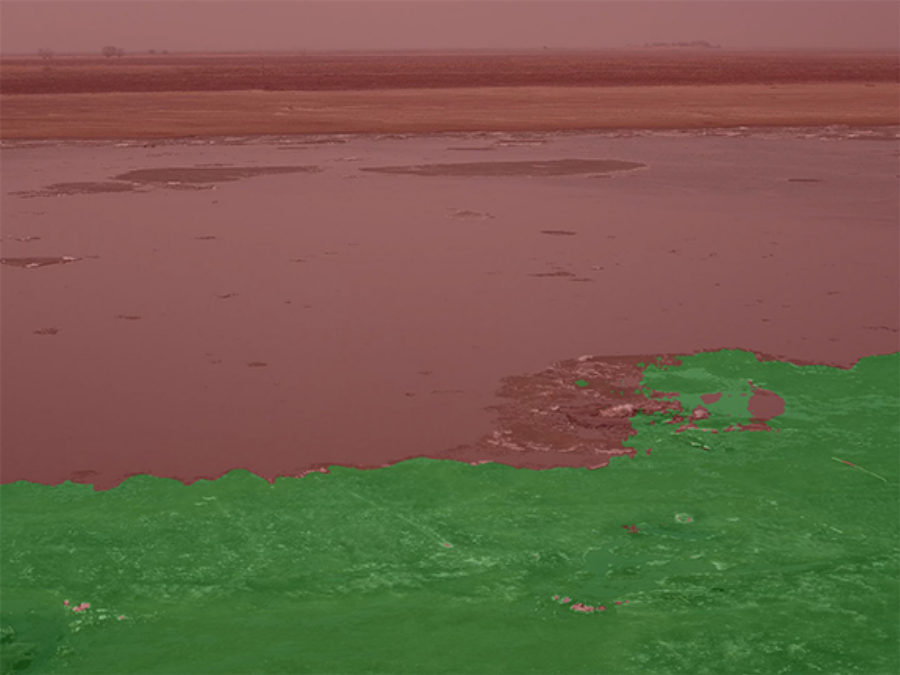}}
				\subfigure[IceHrNet(ours)]{\includegraphics[width=0.24\linewidth]{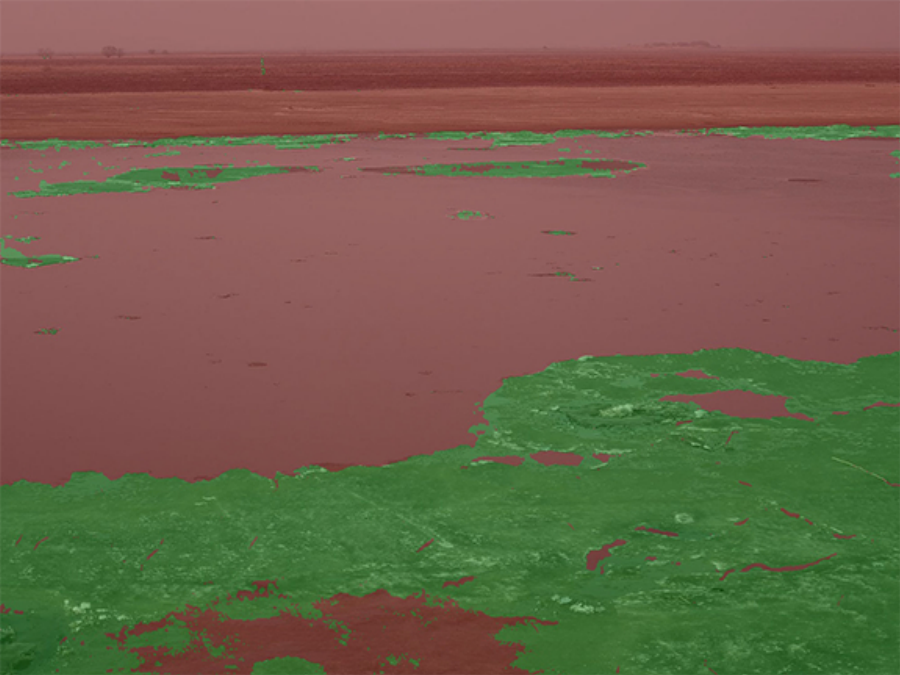}}
			\end{minipage}
			\begin{minipage}[b]{\linewidth}
				\subfigure[UAV Imagery]{\includegraphics[width=0.24\linewidth]{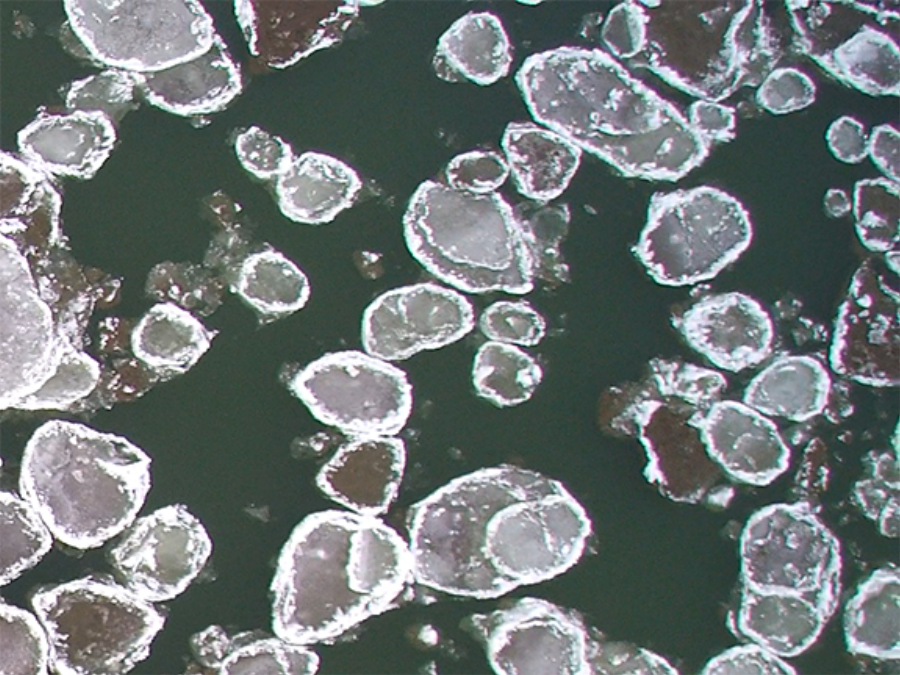}}
				\subfigure[DeepLabV3+]{\includegraphics[width=0.24\linewidth]{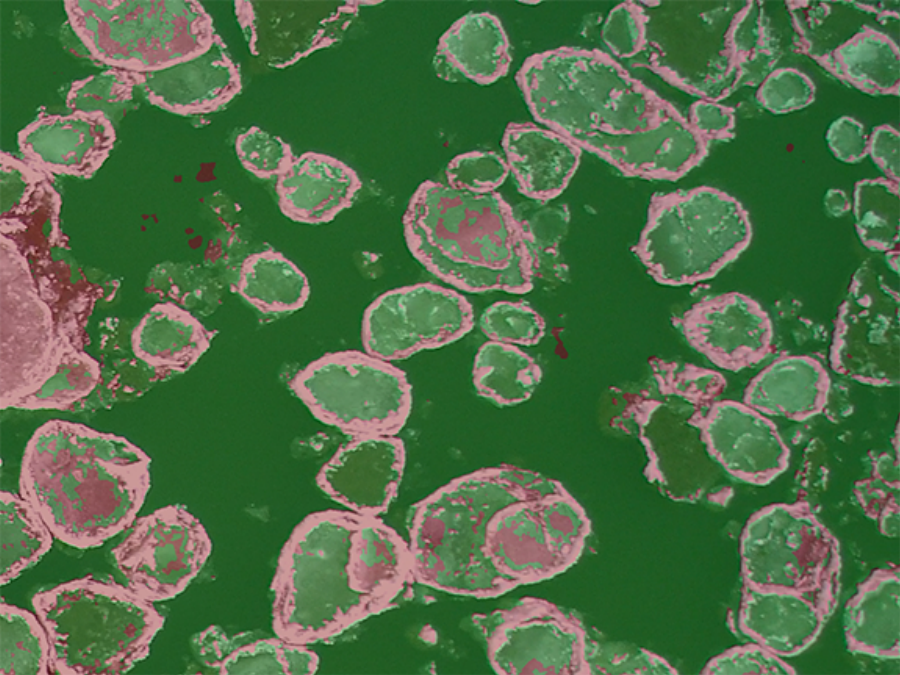}}
				\subfigure[DeepLabV3+]{\includegraphics[width=0.24\linewidth]{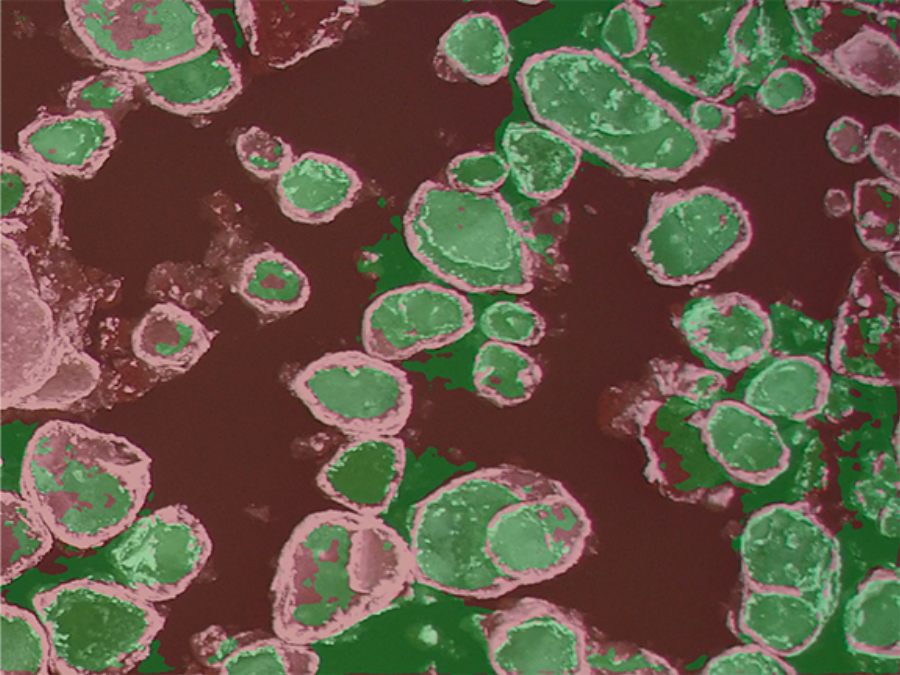}}
				\subfigure[DeepLabV3+]{\includegraphics[width=0.24\linewidth]{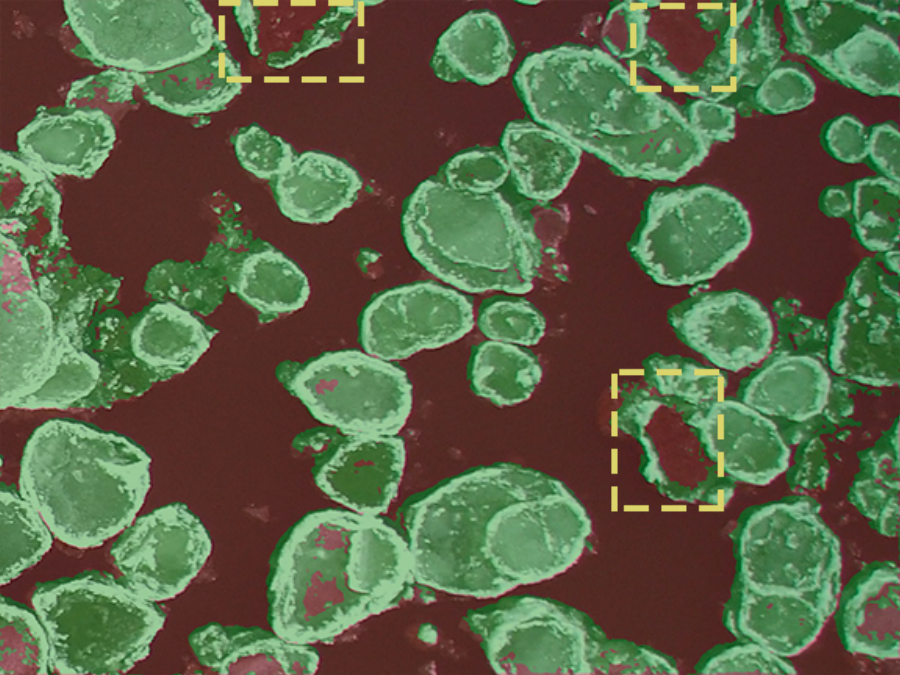}}
			\end{minipage}
			\begin{minipage}[b]{\linewidth}
				\subfigure[]{\includegraphics[width=0.24\linewidth]{blank.png}}
				\subfigure[K-Net]{\includegraphics[width=0.24\linewidth]{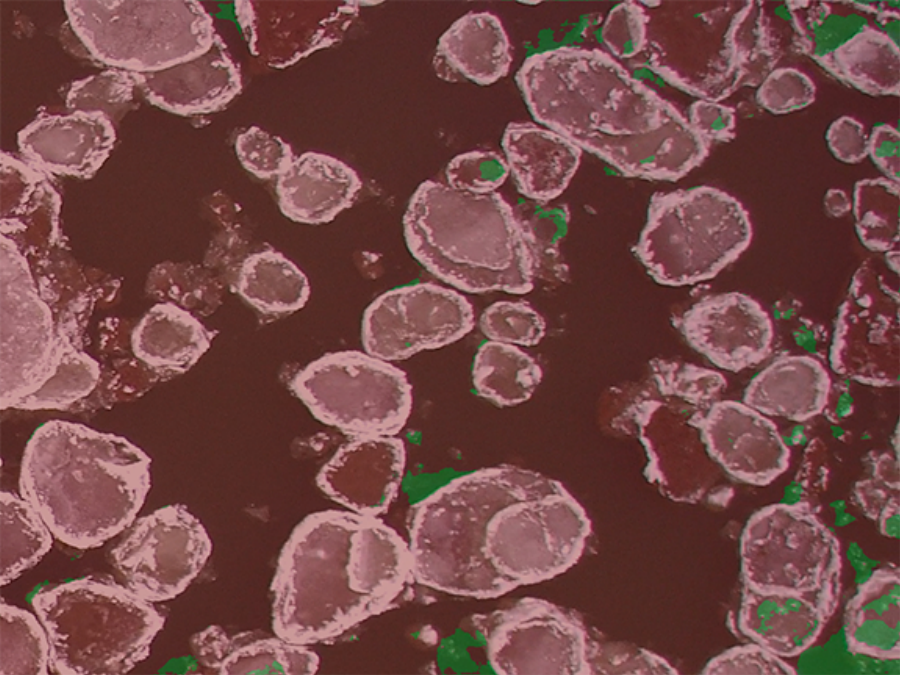}}
				\subfigure[K-Net]{\includegraphics[width=0.24\linewidth]{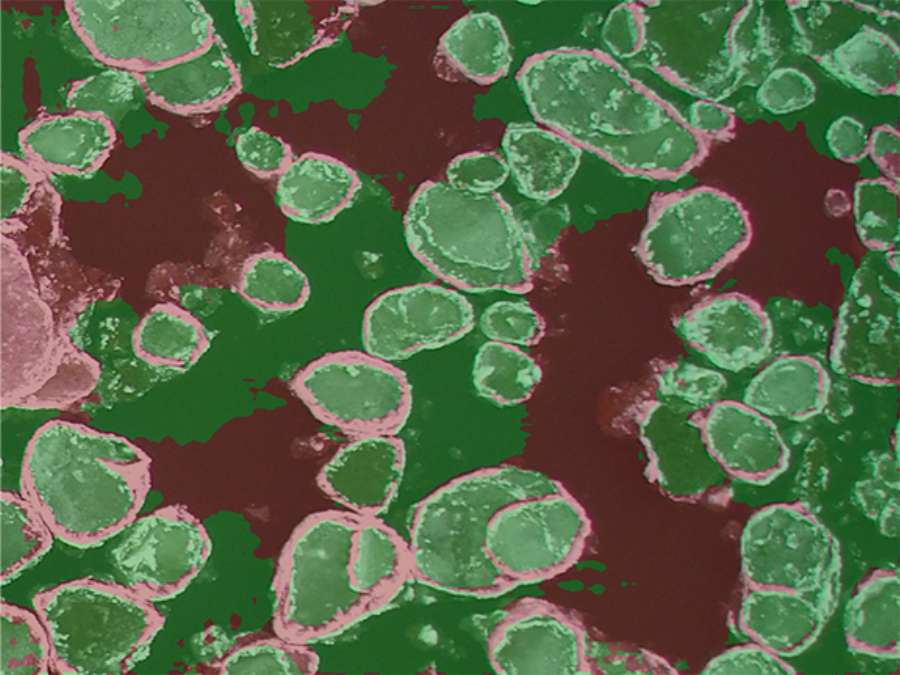}}
				\subfigure[K-Net]{\includegraphics[width=0.24\linewidth]{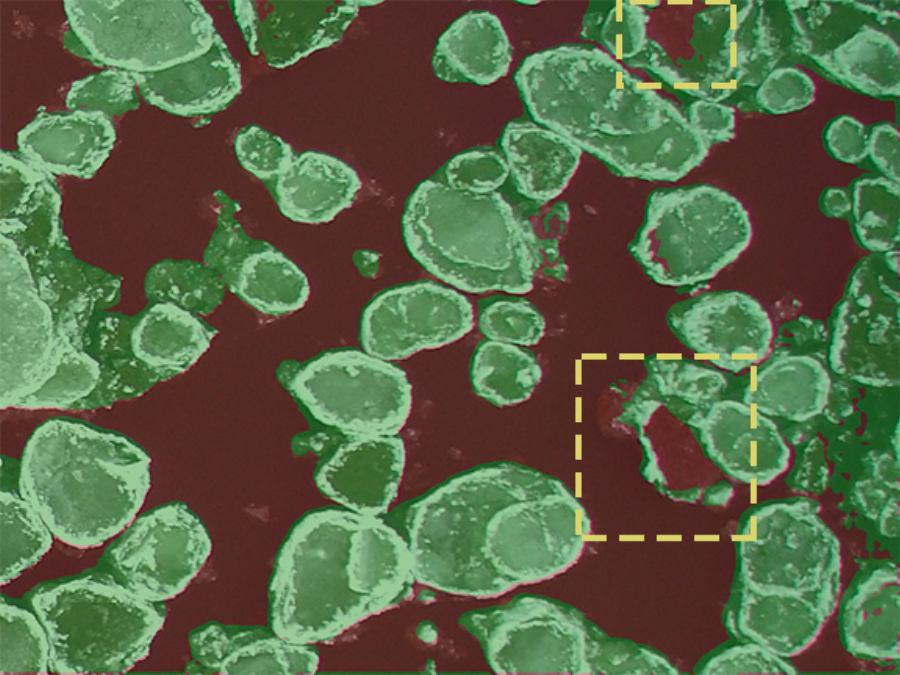}}
			\end{minipage}
			\begin{minipage}[b]{\linewidth}
				\subfigure[]{\includegraphics[width=0.24\linewidth]{blank.png}}
				\subfigure[IceHrNet(ours)]{\includegraphics[width=0.24\linewidth]{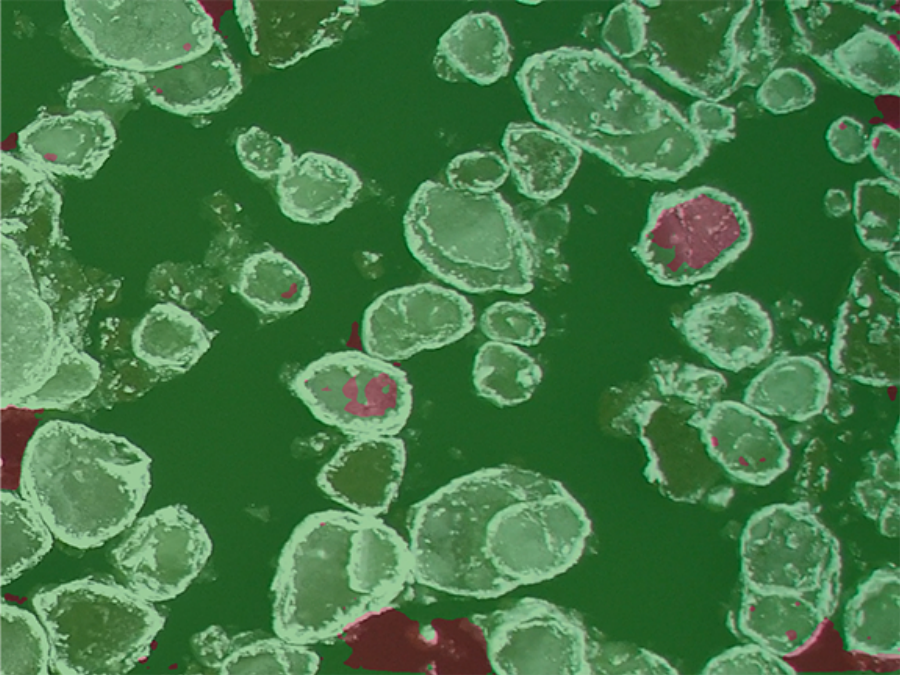}}
				\subfigure[IceHrNet(ours)]{\includegraphics[width=0.24\linewidth]{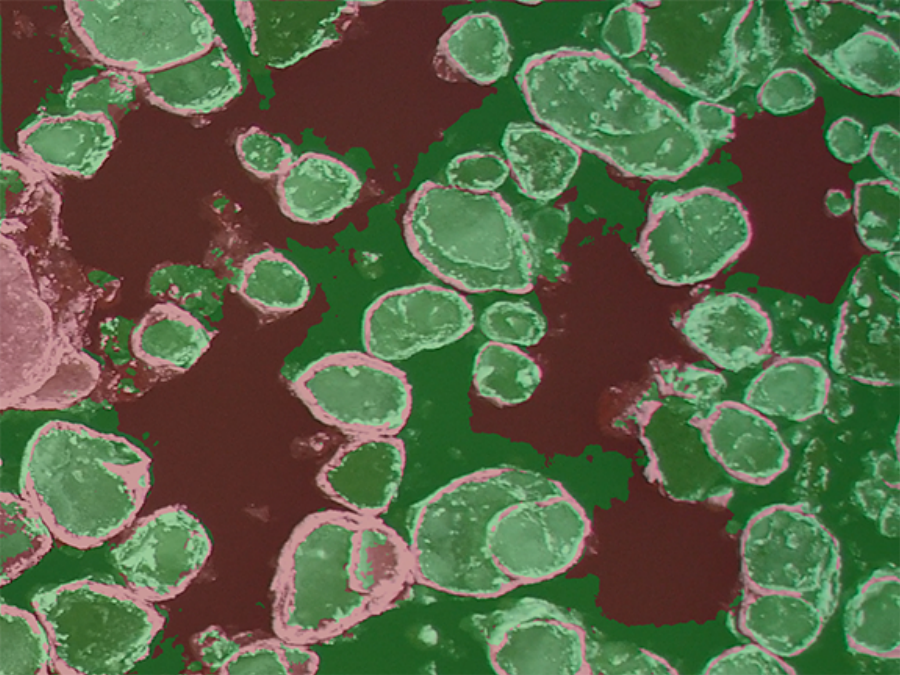}}
				\subfigure[IceHrNet(ours)]{\includegraphics[width=0.24\linewidth]{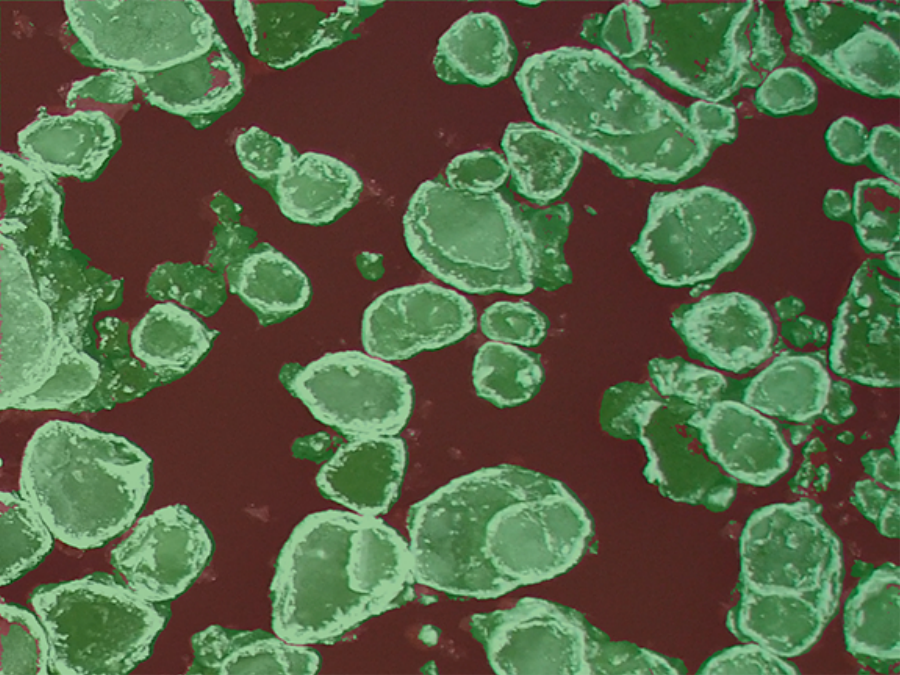}}
			\end{minipage}
		\end{minipage}
	}
	\caption{Visualization results on Zero-Shot Style Transfer Learning. The first original image from Fixed Camera Imagery. The second original image is from UAV Imagery. From the comparison results, it can be seen that our Zero-Shot strategy of Advanced Style Transfer Learning can achieve practical purposes.} \label{Figure11}
\end{figure*}

As can be seen from \textbf{Table \ref{tbl6}} and \textbf{Table \ref{tbl7}}, the segmentation effect of the semantic segmentation network trained by our advanced style transfer learning strategy in the target dataset is close to that of training directly using the target dataset and has achieved optimal results.

Observing \textbf{Table \ref{tbl6}}, in the Fixed Camera Imagery dataset with texture features as the main feature, IceHrNet achieved a score of 0.6 when trained on the source domain dataset that only underwent conventional style transferring, while DeepLabV3+ and KNet were only around 0.3. IceHrNet is twice as much as them. Under the advanced style transfer learning strategy, IceHrNet also achieved an increase of 2 percentage points.

Observing Table \textbf{Table \ref{tbl7}}, DeepLabV3+ and K-Net demonstrated their advantages in the UAV Imagery dataset with texture and shape features as the main features. In the results without style transferring or conventional style transfer learning, they effectively obtained shape feature memory from the source domain dataset and achieved the best results. However, after the advanced style transfer learning strategy, the advantages of IceHrNet were demonstrated and the optimal results were achieved.

For the visualization of the segmentation effect comparison see \textbf{Figure \ref{Figure11}}.

\section{Discussion}

Since the development of neural style transfer, style transfer has naturally been an effective method of transfer learning. However, how to better explore the potential of style transfer in transfer learning is crucial. We analyzed the characteristics of style transfer and proposed a simple strategy to achieve a practical level of style transfer for source domain datasets

After experiments, we found that different networks are sensitive to different features, and thus came up with ideas for further improvements:
\begin{enumerate}
	\item Improve the semantic segmentation network structure to enhance its performance in feature extraction. Networks that are sensitive to multiple features may achieve better results overall.
	\item To improve the strategy of style transfer learning, consider transferring other high-level semantic features while transferring texture features.
\end{enumerate}

The river ice semantic segmentation method using zero-shot style transfer learning can quickly extract river ice pixels and further complete the task of identifying river ice conditions.

\section{Conclusions}

The success of our work is attributed to two aspects: First, we designed an IceHrNet semantic segmentation network that is more sensitive to shallow texture features. The second is an advanced style transfer learning strategy that fully utilizes the texture differences between target domain categories. Coupled with texture-sensitive semantic segmentation networks, it can quickly achieve zero-shot transfer learning across domains. Therefore, the combination of IceHrNet and the advanced style transfer learning strategy proposed by us is very suitable for segmenting scenes where the object is irregularly shaped. With the help of datasets with other similar domains, zero-shot style transfer can be quickly carried out to obtain semantic segmentation applications in the target scenarios.

\textbf{Author Contributions:} Conceptualization, Z.Y., Y.Y., Y.Z. and X.Z.; methodology, Z.Y. and X.Z.; software, Z.Y.; validation, X.L.; investigation, X.C.; resources, J.Z.; data curation, R.T.; writing—original draft, Z.Y.; writing—review and editing, Y.Y.; visualization, R.T.; supervision, Y.Z.; project administration, Y.Y.; funding acquisition, J.Z. All authors have read and agreed to the published version of the manuscript.

\textbf{Funding:} This work was supported by the Central Public-interest Scientific Institution Basal Research Fund [grant number NO.Y522011].

\textbf{Data Availability Statement: }Data used in this study is available upon request to the corresponding author. Code is available at https://github.com/PL23K/IceHrNet accessed on September 25, 2023.

\textbf{Acknowledgments:} Thanks to Singh et al. for sharing his UAV Imagery dataset on the Internet, to the Canadian Ice Service for sharing the precious Arctic Regional Sea Ice Charts. to the Central Public-interest Scientific Institution Basal Research Fund(NO.Y522011) for funding.

\textbf{Conflicts of Interest:} The authors declared no potential conflict of interest with respect to the research, authorship, and/or publication of this article.










\bibliographystyle{cas-model2-names}

\bibliography{cas-refs}



\end{document}